\documentclass[preprint,5p,times,twocolumn]{elsarticle}

\usepackage{amsmath}
\usepackage{amssymb}
\usepackage{balance}
\usepackage{epstopdf}	% convert .eps figures to .pdf
\usepackage[table,xcdraw]{xcolor}
\usepackage{url}
\usepackage{subcaption}
\usepackage{caption}
\usepackage{float}

\usepackage{atbegshi}% http://ctan.org/pkg/atbegshi
\graphicspath{{./figures/}}

\makeatletter
\def\ps@pprintTitle{%
   \let\@oddhead\@empty
   \let\@evenhead\@empty
   \let\@oddfoot\@empty
   \let\@evenfoot\@oddfoot
}
\makeatother

\begin{document}

\title{\LARGE \bf 
Optimal Trajectory Planning for Cinematography\\ with Multiple Unmanned Aerial Vehicles}

\author[1]{Alfonso Alcantara}
\ead{aamarin@us.es}
\author[1]{Jesus Capitan}
\ead{jcapitan@us.es}
\author[2]{Rita Cunha}
\ead{rita@isr.utl.pt}
\author[1]{Anibal Ollero}
\ead{aollero@us.es}

\address[1]{GRVC Robotics Laboratory, University of Seville, Spain.}
\address[2]{Institute for Systems and Robotics, Instituto Superior Técnico, Universidade de Lisboa, Portugal.}

\tnotetext[1]{This work was partially funded by the European Union's Horizon 2020 research and innovation programme under grant agreements No 731667 (MultiDrone), and by the MULTICOP project (Junta de Andalucia, FEDER Programme, US-1265072).}

\begin{abstract}
This paper presents a method for planning optimal trajectories with a team of \emph{Unmanned Aerial Vehicles} (UAVs) performing autonomous cinematography. The method is able to plan trajectories online and in a distributed manner, providing coordination between the UAVs. We propose a novel non-linear formulation for this challenging problem of computing multi-UAV optimal trajectories for cinematography; integrating UAVs dynamics and collision avoidance constraints, together with cinematographic aspects like smoothness, gimbal mechanical limits and mutual camera visibility. We integrate our method within a hardware and software architecture for UAV cinematography that was previously developed within the framework of the \emph{MultiDrone} project; and demonstrate its use with different types of shots filming a moving target outdoors. We provide extensive experimental results both in simulation and field experiments. We analyze the performance of the method and prove that it is able to compute online smooth trajectories, reducing jerky movements and complying with cinematography constraints.  
\end{abstract}

\begin{keyword}
Optimal trajectory planning \sep UAV cinematography \sep Multi-UAV coordination
\end{keyword}

\maketitle

\section{Introduction}

% UAVs are nice for cinematography. 
% Multiple UAVs even more. 
Drones or \emph{Unmanned Aerial Vehicles} (UAVs) are spreading fast for aerial photography and cinematography, mainly due to their maneuverability and their capacity to access complex filming locations in outdoor settings. From the application point of view, UAVs present a remarkable potential to produce unique aerial shots at reduced costs, in contrast with other alternatives like dollies or static cameras. Additionally, the use of teams with multiple UAVs opens even more  the possibilities for cinematography. On the one hand, large-scale events can be addressed by filming multiple action points concurrently or sequentially. On the other hand, the combination of shots with multiple views or different camera motions broadens the artistic alternatives for the \emph{director}.   

% Challenges
Currently, most UAVs in cinematography are operated in manual mode by an expert pilot. Besides, an additional qualified operator is required to control the camera during the flight, as taking aerial shots can be a complex and overloading task. Even so, the manual operation of UAVs for aerial cinematography is still challenging, as multiple aspects need to be considered: performing smooth trajectories to achieve aesthetic videos, tracking actors to be filmed, avoiding collisions with potential obstacles, keeping other cameras out of the field of view, etc. 

% What is out there
There exist commercial products (e.g., \textit{DJI Mavic}~\cite{mavic} or \textit{Skydio}~\cite{skydio}) that cope with some of the aforementioned complexities implementing semi-autonomous functionalities, like \textit{auto-follow} features to track an actor or simplistic collision avoidance. However, they do not address cinematographic principles for multi-UAV teams, as e.g., planning trajectories considering gimbal physical limitations or inter-UAV visibility. Therefore, solutions for autonomous filming with multiple UAVs are of interest. Some authors~\cite{bonatti_jfr20} have shown that planning trajectories ahead several seconds is required in order to fulfill with cinematographic constraints smoothly. Others~\cite{naegeli_tg17,galvane_tg18} have even explored the multi-UAV problem, but online trajectory planning for multi-UAV cinematography outdoors is still an open issue.  

% What we propose
In this paper, we propose a method for online planning and execution of trajectories with a team of UAVs taking cinematography shots. We develop an optimization-based technique that runs on the UAVs in a distributed fashion, taking care of the control of the UAV and the gimbal motion simultaneously. Our method aims at providing smooth trajectories for visually pleasant video output; integrating cinematographic constraints imposed by the shot types, the gimbal physical limits, the mutual visibility between cameras and the avoidance of collisions. 

\begin{figure}[htb]
	\centering
	\includegraphics[width=\linewidth]{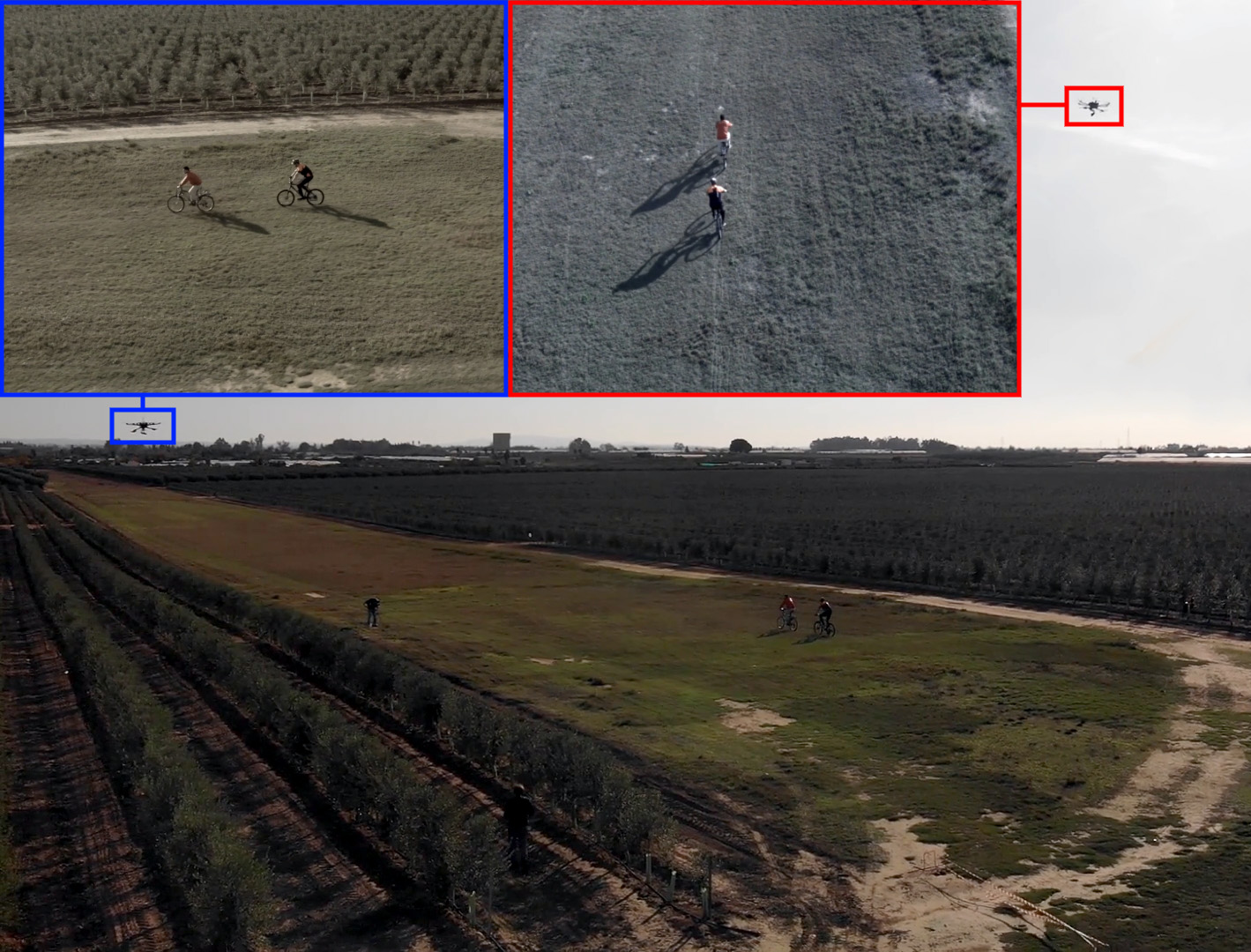}
	\caption{Cinematography application with two UAVs filming a cycling event. Bottom, aerial view of the experiment with two moving cyclists. Top, images taken from the cameras on board each UAV. }
	\label{fig:MissionOf3Drones}
\end{figure}

% Mention the director, dashboard or planner
This work has been developed within the framework of the EU-funded project \textit{MultiDrone}\footnote{\url{https://multidrone.eu}.}, whose objective was to create a complete system for autonomous cinematography with multiples UAVs in outdoor sport events (see Figure~\ref{fig:MissionOf3Drones}).
MultiDrone addressed different aspects to build a complete architecture: a set of high-level tools so that the cinematography director can define shots for the mission~\cite{montes_appsci20}; planning methods to assign and schedule the shots among the UAVs efficiently and considering battery constraints~\cite{caraballo_iros20}; vision-based algorithms for target tracking on the camera image~\cite{nousi_icip19}, etc.
In this paper, we focus on the autonomous execution of shots with a multi-UAV team. 
%Given a taxonomy of representative shots for UAV cinematography~\cite{mademlis_tb19,mademlis_cs19}, 
We assume that the director has designed a mission with several shots; and that there is a planning module that has assigned a specific shot to each UAV. Then, our objective is to plan trajectories in order to execute all shots online in a coordinated manner.  

%%%%%%%%%%%%%%%%%%%%%%%%%%%%%%%%%%%%%%%%%%%%%
\subsection{Related work}
\label{sec:soa}

% Optimal trajectory planning without cinematography
Optimal trajectory planning for UAVs is a commonplace problem in the robotics community. A typical approach is to use optimization-based techniques to generate trajectories from polynomial curves minimizing their derivative terms for smoothness, e.g., the fourth derivative or snap~\cite{richter_isrr16,mellinger_icra11}. This polynomial trajectories have also been applied to optimization problems with multiple UAVs~\cite{turpin_auro14}. \emph{Model Predictive Control} (MPC) is another widespread technique for optimal trajectory planning \cite{baca_iros18}, a dynamic model of the UAV is used for predicting and optimizing trajectories ahead within a receding horizon. Some authors~\cite{kamel_iros17} have also used MPC-based approaches for multi-UAV trajectory planning with collision avoidance and non-linear models. In the context of multi-UAV target tracking, others~\cite{price_ral18,Tallamraju2018} have combined MPC with potential fields to address the non-convexity induced by collision avoidance constraints.   
In~\cite{kratky_ral20}, a constrained optimization problem is formulated to maintain a formation where a leader UAV takes pictures for inspection in dark spaces, while others illuminate the target spot supporting the task. The system is used for aerial documentation within historical buildings.  

% Target tracking not considering cinematography rules
Additionally, there are works in the literature just for target tracking with UAVs, proposing alternative control techniques like classic PID~\cite{teuliere_iros11} or LQR~\cite{naseer_iros13} controllers. The main issues with all these methods for trajectory planning and target tracking are that they either do not consider cinematographic aspects explicitly or do not plan ahead in time for horizons long enough.

% In~\cite{kamel_iros17}, they assign a hierarchical scheme where the first drone has the top priority and do not have to avoid anyone, the second one has to avoid the first one, the third one has to avoid the first two, etc. 
% In~\cite{price_ral18}, Model Predictive Control (MPC) is enhanced with obstacle avoidance based on potential fields for outdoor target tracking. They successfully perform target tracking although they do not comply with cinematography rules.
% For instance,~\cite{Tallamraju2018} calculates a geometry formation over the target using a high-level planner and then uses potential field forces to maintain the distance with respect to other drones in a decentralized way. 

% Computer graphics
In the computer animation community, there are several works related with trajectory planning for the motion of virtual cameras \cite{Christie2008}. They typically use offline optimization to generate smooth trajectories that are visually pleasant and comply with certain cinematographic aspects, like the \emph{rule of thirds}.
However, many of them do not ensure physical feasibility to comply with UAV dynamic constraints and they assume full knowledge of the environment map. 
In terms of optimization functions, several works consider similar terms to achieve smoothness. For instance, authors in~\cite{joubert_siggraph15} model trajectories as polynomial curves whose coefficients are computed to minimize snap (fourth derivative). They also check dynamic feasibility along the planned trajectories, and the user is allowed to adjust the UAV velocity at execution time. A similar application to design UAV trajectories for outdoor filming is proposed in~\cite{joubert_arxiv16}. Timed reference trajectories are generated from 3D positions specified by the user, and the final timing of the shots is addressed designing easing curves that drive the UAV along the planned trajectory (i.e., curves that modify the UAV velocity profile). In~\cite{gebhardt_chi16}, aesthetically pleasant footage is achieved by penalizing the snap of the UAV trajectory and the jerk (third derivative) of the camera motion. An iterative quadratic optimization problem is formulated to compute trajectories for the camera and the look-at point (i.e., the place where the camera is pointing at). They also include collision avoidance constraints, but the method is only tested indoors. 

Although these articles on computer graphics approach the problem mainly through offline optimization, some of them have proposed options to achieve real-time performance, like planning in a \emph{toric space}~\cite{lino_tog15} or interpolating polynomial curves~\cite{galvane_eurographics16, joubert_arxiv16}. In general, these works present interesting theoretical properties, but they are restricted to offline optimization with a fully known map of the scenario and static or close-to-static guided tour scenes, i.e., without moving actors. 

\begin{table*}[]
\centering
{%
\begin{tabular}{|c|c|c|c|c|c|c|c|}
\hline
References & Online & Scene & \begin{tabular}[c]{@{}c@{}}UAVs \\ Dynamics\end{tabular} & \begin{tabular}[c]{@{}c@{}}Collision\\ Avoidance\end{tabular} & \begin{tabular}[c]{@{}c@{}}Mutual\\ Visibility\end{tabular} & Outdoors & Multiples UAVs \\ \hline \hline
{[}23{]}   & No              & Static         & No                                                                & No                                                                     & No                                                                   & No                & No                      \\ \hline
{[}20{]}   & No              & Static         & Yes                                                               & No                                                                     & No                                                                   & Yes               & No                      \\ \hline 
2

{[}22{]}   & No              & Static         & Yes                                                               & Yes                                                                    & No                                                                   & No                & No                      \\ \hline
{[}21{]}   & No              & Static         & Yes                                                               & Actor                                                                  & No                                                                   & Yes               & No                      \\ \hline
{[}24{]}   & No              & Dynamic        & No                                                                & No                                                                     & No                                                                   & No                & No                      \\ \hline
{[}27{]}   & Yes             & Dynamic        & Yes                                                               & Yes                                                                    & No                                                                  & No                & No                      \\ \hline
{[}4{]}    & Yes             & Dynamic        & Yes                                                               & Actor                                                                  & Yes                                                                  & No                & Yes                     \\ \hline
{[}25{]}   & Yes             & Dynamic        & Yes                                                               & Actor                                                                  & No                                                                   & Yes               & No                      \\ \hline
{[}5{]}    & Yes             & Dynamic        & Yes                                                               & Yes                                                                    & Yes                                                                  & No                & Yes                     \\ \hline
{[}26{]}   & Yes             & Dynamic        & Yes                                                               & Yes                                                                    & No                                                                  & Yes               & No                      \\ \hline
{[}3{]}    & Yes             & Dynamic        & Yes                                                               & Yes                                                                    & No                                                                  & Yes               & No                      \\ \hline
Ours       & Yes             & Dynamic        & Yes                                                               & Yes                                                                    & Yes                                                                  & Yes               & Yes                     \\ \hline
\end{tabular}%
}
\caption{Related works on trajectory planning for UAV cinematography. We indicate whether computation is online or not, the type of scene and constraints they consider, and their capacity to handle outdoor applications and multiple UAVs.}
\label{tab:related_work}
\end{table*}

% \begin{figure*}[tb!]
%     \centering
%     \includegraphics[width=2.1\columnwidth]{figures/soa_aerial_cinematography.pdf}
%     \caption{Related works on trajectory planning for UAV cinematography. We indicate whether computation is online or not, the type of scene and constraints they consider, and their capacity to handle outdoor applications and multiples UAVs.}
%     \label{fig:related_work}
% \end{figure*}

% Cinematographic, outdoors
In the robotics literature, there are works focusing more on filming dynamic scenes and complying with physical UAV constraints. For example, authors in~\cite{Huang} propose to detect limbs movement of a human for outdoor filming. Trajectory planning is performed online with polynomial curves that minimize snap. In~\cite{Bonatti2019,bonatti_jfr20}, they present an integrated system for outdoor cinematography, combining vision-based target localization with trajectory planning and collision avoidance. For optimal trajectory planning, they apply gradient descent with differentiable cost functions. Smoothness is achieved minimizing trajectory jerk; and shot quality by defining objective curves fulfilling cinematographic constraints associated with relative angles w.r.t. the actor and shot scale. Cinematography optimal trajectories have also been computed in real time through receding horizon with non-linear constraints~\cite{naegeli_ral17}. The user inputs framing objectives for one or several targets on the image, and errors of the image target projections, sizes and relative viewing angles are minimized; satisfying collision avoidance constraints and target visibility. The method behaves well for online numerical optimization, but it is only tested in indoor settings.     

% Learning how to achieve aesthetically better shots, which is not our focus, outdoors.
Some of the aforementioned authors from robotics have also approached UAV cinematography applying machine learning techniques. In particular, learning from demonstration to imitate professional cameraman's behaviors~\cite{huang_icra19} or reinforcement learning to achieve visually pleasant shots~\cite{Gschwindt2019}. In general, most of these cited works on robotics present results quite interesting in terms of operation outdoors or online trajectory planning, but they always restrict to a single UAV. 

% Multiples drones with cinematography
Regarding methods for multiple UAVs, there is some related work which is worth mentioning. In~\cite{naegeli_tg17}, a non-linear optimization problem is solved in a receding horizon fashion, taking into account collision avoidance constraints with the filmed actors and between the UAVs. Aesthetic objectives are introduced by the user as virtual reference trails. Then, UAVs receive current plans from all others at each planning iteration and compute collision-free trajectories sequentially. A UAV toric space is proposed in~\cite{galvane_tg18} to ensure that cinematographic properties and dynamic constraints are ensured along the trajectories. Non-linear optimization is applied to generate polynomial curves with minimum curvature variation, accounting for target visibility and collision avoidance. The motion of multiple UAVs around dynamic targets is coordinated by means of a centralized master-slave approach to solve conflicts. Even though these works present promising results for multi-UAV teams, they are only demonstrated at indoor scenarios where a \emph{Vicon} motion capture system provides accurate positioning for all targets and UAVs. 
These works present quite valuable contributions for cinematography with multiple UAVs, but they are evaluated in indoor settings. The specifics of the outdoor scenarios considered in our work are different in several aspects, as the environment is less controlled: UAVs require more payload to carry onboard cameras with better lenses and equipment for larger range communication; achieving smooth trajectories is more complex due to external factors such as wind gusts or communication delays; UAV positioning is less accurate in general; and so on.

To sum up, Table~\ref{tab:related_work} shows the main related works on trajectory planning for UAV cinematography and their corresponding properties. We indicate whether computation is online or offline, whether the scene contains dynamic targets to be filmed and whether UAV dynamics are included as constraints. We also analyze the type of collision avoidance: none (\emph{No}), with the actor being filmed (\emph{Actor}) or with external obstacles and other UAVs (\emph{Yes}). Works which address mutual visibility constraints between multiple cameras are mentioned specifically. Finally, we indicate whether each method includes evaluation in outdoor settings and whether it can handle multiple UAVs.

%%%%%%%%%%%%%%%%%%%%%%%%%%%%%%%%%%%%%%%
\subsection{Contributions}

We propose a novel method to plan online optimal trajectories for a set of UAVs executing cinematography shots. The optimization is performed in a distributed manner, and it aims for smooth trajectories complying with dynamic and cinematographic constraints. We extend our previous work~\cite{ecmr19} in optimal trajectory planning for UAV cinematography as follows: (i) we cope with multiple UAVs integrating new constraints for inter-UAV collisions and mutual visibility; (ii) we present additional simulation results to evaluate the method with different types of shots; and (iii) we demonstrate the system in field experiments with multiple UAVs filming dynamic scenes. Therefore, the main novelty of our method is the multi-UAV coordination to combine the execution of several types of shots simultaneously in outdoor scenarios, with the specific challenges that those environments involve. More particularly, our main contributions are the following:
 
\begin{itemize}
    \item We propose a novel formulation of the trajectory planning problem for UAV cinematography. We model both UAV and gimbal motion (Section~\ref{sec:model}), but decouple their control actions.
    \item We propose a non-linear, optimization-based method for trajectory planning (Section~\ref{sec:method}). Using a receding horizon scheme, trajectories are planned and executed in a distributed manner by a team of UAVs providing multiple views of the same scene. The method considers UAV dynamic constraints, and imposes them to avoid predefined no-fly zones or collisions with others. Cinematographic aspects imposed by shot definition, camera mutual visibility and gimbal physical bounds are also addressed.  
    Trajectories smoothing UAV and gimbal motion are generated to achieve aesthetic video footage.
    \item We describe the complete system architecture on board each UAV and the different types of shot considered (Section~\ref{sec:system_architecture}). The architecture integrates target tracking with trajectory planning and it is such that different UAVs can be executing different types of shot simultaneously. 
    \item We present extensive experimental results (Section~\ref{sec:simulations}) to evaluate the performance of our method for different types of shot. We prove that our method is able to compute smooth trajectories reducing jerky movements in real time, and complying with the cinematographic restrictions. Then, we demonstrate our system in field experiments with three UAVs planning trajectories online to film a moving actor (Section~\ref{sec:fieldExp}). 
\end{itemize}

%%%%%%%%%%%%%%%%%%%%%%%%%%%%%%%%%%%%%%%%%%%%%%%%%%%%%%
%%%%%%%%%%%%%%%%%%%%%%%%%%%%%%%%%%%%%%%%%%%%%%%%%%%%%%

\section{Dynamic Models}
\label{sec:model}

This section presents our dynamic models for UAV cinematographers. We model the UAV as a quadrotor with a camera mounted on a gimbal of two degrees of freedom.  

\subsection{UAV model}

\begin{figure}
    \centering
    \includegraphics[width=1\columnwidth]{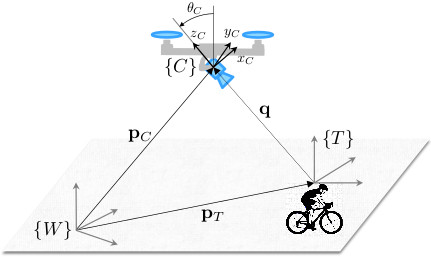}
    \caption{Definition of reference frames used. The origins of the camera and quadrotor frames coincide. The camera points to the target.}
    \label{fig:notation}
\end{figure}

Let $\{W\}$ denote the world reference frame with origin fixed in the environment and East-North-Up (ENU) orientation. Consider also three additional reference frames (see Figure~\ref{fig:notation}): the quadrotor reference frame $\{Q\}$ attached to the UAV with origin at the center of mass, the camera reference frame $\{C\}$ with $z$-axis aligned with the optical axis but with opposite sign, and the target reference frame $\{T\}$ attached to the moving target that is being filmed. For simplicity, we assumed that the origins of $\{Q\}$ and $\{C\}$ coincide. 

The configuration of $\{Q\}$ with respect to $\{W\}$ is denoted by $(\mathbf{p}_Q,R_Q)\in\mathbb{SE}(3)$, where $\mathbf{p}_Q\in\mathbb{R}^3$ is the position of the origin of $\{Q\}$ expressed in $\{W\}$ and $R_Q\in\mathbb{SO}(3)$ is the rotation matrix from $\{Q\}$ to $\{W\}$. Similarly, the configurations of $\{T\}$ and $\{C\}$ with respect to $\{W\}$ are denoted by $(\mathbf{p}_T,R_T)\in\mathbb{SE}(3)$ and $(\mathbf{p}_C,R_C)\in\mathbb{SE}(3)$, respectively.

We model the quadrotor dynamics as a linear double integrator model:
\begin{align}
    \dot{\mathbf{p}}_Q &= \mathbf{v}_Q \nonumber\\
    \dot{\mathbf{v}}_Q &= \mathbf{a}_Q,
\label{eq:dynamics}
\end{align}
\noindent where $\mathbf{v}_Q = [v_x\ v_y\ v_z]^T\in\mathbb{R}^3$ is the linear velocity and $\mathbf{a}_Q = [a_x\ a_y\ a_z]^T \in\mathbb{R}^3$ is the linear acceleration. We assume that the linear acceleration  $\mathbf{a}_Q$ takes the form:
\begin{equation}
    \mathbf{a}_Q =  -g\mathbf{e}_3 +  R_Q \frac{T}{m} \mathbf{e}_3,
    \label{eq:acceleration_model}
\end{equation}
where $m$ is the quadrotor mass, $g$ the gravitational acceleration, $T\in\mathbb{R}$ the scalar thrust, and $\mathbf{e}_3 = [0\ 0\ 1]^T$. 

For the sake of simplicity, we use the 3D acceleration $\mathbf{a}_Q$ as control input; although the thrust $T$ and rotation matrix $R_Q$ could also be recovered from 3D velocities and accelerations. %For that, we first parameterize $R_Q$ by the $Z$-$Y$-$X$ Euler angles $\boldsymbol{\lambda}_Q = [ \phi_Q,\theta_Q,\psi_Q]^T$, such that 
%\begin{equation}
%    R_Q = R_z(\psi_Q) R_y(\theta_Q) R_x(\phi_Q).
%\end{equation}
If we restrict the yaw angle $\psi_Q$ to keep the quadrotor's front pointing forward in the direction of motion such that:
\begin{equation}
    \psi_Q = \mathrm{atan2}(v_y,v_x),
\end{equation}
%which yields
%\begin{align}
%    R_z(-\psi_Q) \mathbf{v}_Q &= 
%     \begin{bmatrix}
%        \sqrt{v_x^2+v_y^2} \\
%        0\\
%        v_z
%    \end{bmatrix}.
%\end{align}
%Finally, to recover the thrust $T$, the roll angle $\phi_Q$, and the pitch angle $\theta_Q$, we use \eqref{eq:acceleration_model} and apply some algebraic manipulations to obtain
%\begin{align}
%    \frac{T}{m}R_y(\theta_Q)R_x(\phi_Q)\mathbf{e}_3 &= R_z(-\psi_Q) (\mathbf{a}_Q+g\mathbf{e}_3) \nonumber\\
%    \frac{T}{m}\begin{bmatrix}
%\cos(\phi_Q)\sin(\theta_Q)\\
%           -\sin(\phi_Q)\\
% \cos(\phi_Q)\cos(\theta_Q)
%    \end{bmatrix} &= \begin{bmatrix} a_x\cos(\psi_Q) + a_y\sin(\psi_Q)\\
% a_y\cos(\psi_Q) - a_x\sin(\psi_Q)\\
%                    a_z + g
%    \end{bmatrix}.
%\end{align}
\noindent then the thrust $T$ and the $Z$-$Y$-$X$ Euler angles $\boldsymbol{\lambda}_Q = [ \phi_Q,\theta_Q,\psi_Q]^T$ can be obtained from $\mathbf{v}_Q$ and $\mathbf{a}_Q$ according to:
\begin{equation}\label{eq:Tlambda_Q}
    \begin{cases}
    T = m \|\mathbf{a}_Q+g\mathbf{e}_3\|\\
    \psi_Q = \mathrm{atan2}(v_y,v_x)\\
    \phi_Q = -\arcsin((a_y\cos(\psi_Q) - a_x\sin(\psi_Q))/\|\mathbf{a}_Q+g\mathbf{e}_3\|)\\
    \theta_Q = \mathrm{atan2}(a_x\cos(\psi_Q) + a_y\sin(\psi_Q),a_z + g)
    \end{cases}
\end{equation}

\subsection{Gimbal angles}

Let $\boldsymbol{\lambda}_C = [ \phi_C,\theta_C,\psi_C]^T$ denote
the $Z$-$Y$-$X$ Euler angles that parametrize the rotation matrix $R_C$, such that:
\begin{equation}
    R_C = R_z(\psi_C) R_y(\theta_C) R_x(\phi_C).
\end{equation}

In our system, we decouple gimbal motion with an independent gimbal attitude controller that ensures that the camera is always pointing towards the target during the shot, as in~\cite{bonatti_jfr20}. This reduces the complexity of the planning problem and allows us to control the camera based on local perception feedback if available, accumulating less errors. We also consider that the time-scale separation between the "faster" gimbal dynamics and "slower" quadrotor dynamics is sufficiently large to neglect the gimbal dynamics and assume an exact match between the desired and actual orientations of the gimbal. 
In order to define $R_C$, let us introduce the relative position: 
\begin{equation}
 \mathbf{q} = \begin{bmatrix}
        q_x & q_y & q_z
    \end{bmatrix}^T
= \mathbf{p}_C - \mathbf{p}_T,   
\end{equation}

\noindent and assume that the UAV is always above the target, i.e., $q_z>0$, and not directly above the target, i.e., $[q_x\ q_y]\neq 0$. Then, the gimbal orientation $R_C$ that guarantees that the camera is aligned with the horizontal plane and pointing towards the target is given by:
\begin{align}
    R_C &= \begin{bmatrix}
-\dfrac{\mathbf{q}\times\mathbf{q}\times \mathbf{e}_3}{\|\mathbf{q}\times\mathbf{q}\times \mathbf{e}_3\|} &    \dfrac{\mathbf{q}\times \mathbf{e}_3}{\|\mathbf{q}\times \mathbf{e}_3\|} & \dfrac{\mathbf{q}}{\|\mathbf{q}\|}
    \end{bmatrix}\nonumber\\
%    &= \begin{bmatrix}
%    * & \frac{q_y}{\sqrt{q_x^2+q_y^2}} & \frac{q_x}{\sqrt{q_x^2+q_y^2+q_z^2}}\\
%    * & \frac{-q_x}{\sqrt{q_x^2+q_y^2}} & \frac{q_y}{\sqrt{q_x^2+q_y^2+q_z^2}}\\
%    \frac{-\sqrt{q_x^2+q_y^2}}{\sqrt{q_x^2+q_y^2+q_z^2}} & 0 & \frac{q_z}{\sqrt{q_x^2+q_y^2+q_z^2}}
&= \begin{bmatrix}
    * & \frac{q_y}{\sqrt{q_x^2+q_y^2}} & *\\
    * & \frac{-q_x}{\sqrt{q_x^2+q_y^2}} & *\\
    \frac{\sqrt{q_x^2+q_y^2}}{\sqrt{q_x^2+q_y^2+q_z^2}} & 0 & \frac{q_z}{\sqrt{q_x^2+q_y^2+q_z^2}}
    \end{bmatrix}.
\end{align}

To recover the Euler angles from the above expression of $R_C$, note that if the camera is aligned with the horizontal plane, then there is no roll angle, i.e. $\phi_C = 0$, and $R_C$ takes the form: 
\begin{equation}\label{eq:RC}
R_C = \begin{bmatrix} 
 \cos(\psi_C)\cos(\theta_C) & -\sin(\psi_C) & \cos(\psi_C)\sin(\theta_C)\\
\cos(\theta_C)\sin(\psi_C) & \cos(\psi_C) & \sin(\psi_C)\sin(\theta_C) \\
-\sin(\theta_C) &0 & \cos(\theta_C)
    \end{bmatrix},
\end{equation}
and we obtain:
\begin{equation}\label{eq:lambda_C}
    \begin{cases}
    \phi_C = 0\\
    \theta_C = \mathrm{atan2}(-\sqrt{q_x^2+q_y^2},q_z)\\
    \psi_C = \mathrm{atan2}(-q_y,-q_x)
    \end{cases}
    \end{equation}
    
%Consequently, the orientation of the gimbal with respect to the quadrotor ${^Q_C}R$ is also determined by these variables, given that
%\begin{align}
%    {^Q_C}R &= (R_Q)^T R_C\nonumber\\
%    &= R_x(-\phi_Q) R_y(-\theta_Q) R_z(\psi_C-\psi_Q) R_y(\theta_C) R_x(\phi_C).
%\end{align}

Our cinematography system is designed to perform smooth trajectories as 
the UAVs are taking their shots, and then using more aggressive maneuvers only to fly between shots without filming. If UAVs fly smoothly, we can assume that their accelerations $a_x$ and $a_y$ are small, and hence, by direct application of Eq.~\eqref{eq:Tlambda_Q}, that their roll and pitch angles are small and $R_x(\phi_Q) \approx R_y(\theta_Q) \approx I_3$. This assumption is relevant to alleviate the non-linearity of the model and achieve real-time numerical optimization. Moreover, it is reasonable during shot execution, as our trajectory planner will minimize explicitly UAV accelerations, and will limit both UAV velocities and accelerations.  

Under this assumption, the orientation matrix of the gimbal with respect to the quadrotor $ {^Q_C}R$ can be approximated by:
\begin{align}
    {^Q_C}R &= (R_Q)^T R_C\nonumber\\
    &\approx   R_z(\psi_C-\psi_Q)R_y(\theta_C)R_x(\phi_C),
\end{align}
and the relative Euler angles $^Q\boldsymbol{\lambda}_C$ (roll, pitch and yaw) of the gimbal with respect to the quadrotor are obtained as:
\begin{equation}\label{eq:Qlambda_C}
    \begin{cases}
    ^Q\phi_C = \phi_C = 0\\
    ^Q\theta_C =\theta_C =  \mathrm{atan2}(-\sqrt{q_x^2+q_y^2},q_z)\\
    ^Q\psi_C = \psi_C -\psi_Q= \mathrm{atan2}(-q_y,-q_x) - \mathrm{atan2}(v_y,v_x)
    \end{cases}
\end{equation}

According to Eq.~\eqref{eq:Tlambda_Q}, \eqref{eq:lambda_C} and \eqref{eq:Qlambda_C}, $\boldsymbol{\lambda}_Q$, $\boldsymbol{\lambda}_C$ and $^Q\boldsymbol{\lambda}_C$ are completely defined by the trajectories of the quadrotor and the target, as explicit functions of $\mathbf{q}$, $\mathbf{v}_Q$, and $\mathbf{a}_Q$.

%%%%%%%%%%%%%%%%%%%%% DERIVATIVES %%%%%%%%%%%%%%%%%%%%%%%%%%%
%%%%%%%%%%%%%%%%%%%%%%%%%%%%%%%%%%%%%%%%%%%%%%%%%%%%%%%%%%%%
% For optimization purposes, it is also convenient to determine the time derivatives of these variables. Noting that
% \begin{equation}
%     \frac{d}{dt} \mathrm{atan2}(f(t), g(t)) = \frac{\dot{f}(t) g(t)- \dot{g}(t) f(t)}
%     {f(t)^2 + g(t)^2},
% \end{equation}
% we can write
% \begin{align}
%     ^Q\dot{\theta}_C &=\dot{\theta}_C = \dfrac{\dot{q}_z\sqrt{q_x^2+q_y^2}}{(q_x^2+q_y^2+q_z^2)}- \dfrac{(q_x\dot{q}_x+q_y\dot{q}_y)q_z}{(q_x^2+q_y^2+q_z^2)\sqrt{q_x^2+q_y^2}}\\
%     ^Q\dot{\psi}_C &= \dot{\psi}_C -\dot{\psi}_Q = \dfrac{\dot{q}_x q_y -\dot{q}_y q_x }{q_x^2+q_y^2} - \dfrac{\dot{v}_y v_x - \dot{v}_x v_y}{v_x^2+v_y^2} 
% \end{align}
% Finally, substituting the expressions for $\dot{q}_x$, $\dot{q}_y$, $\dot{v}_x$, and $\dot{v}_y$, we obtain
% \begin{align}
%     ^Q\dot{\theta}_C &=\dot{\theta}_C = \dfrac{(v_z-v_{zT})\sqrt{q_x^2+q_y^2}}{(q_x^2+q_y^2+q_z^2)}- \dfrac{q_xq_z(v_x-v_{xT})+q_yq_z(v_y-v_{yT})}{(q_x^2+q_y^2+q_z^2)\sqrt{q_x^2+q_y^2}} \\
%     ^Q\dot{\psi}_C &= \dot{\psi}_C -\dot{\psi}_Q = \dfrac{(v_x-v_{xT}) q_y -(v_y-v_{yT}) q_x }{q_x^2+q_y^2} - \dfrac{a_y v_x - a_x v_y}{v_x^2+v_y^2}
% \end{align}

%%%%%%%%%%%%%%%%%%%%%%%%%%%%%%%%%%%%%%%%%%%%%%%%%
%%%%%%%%%%%%%%%%%%%%%%%%%%%%%%%%%%%%%%%%%%%%%%%%%
\section{Optimal Trajectory Planning}
\label{sec:method}

In this section, we describe our method for optimal trajectory planning. We explain how trajectories are computed online in a receding horizon scheme, considering dynamic and cinematographic constraints; and then, how the coordination between multiple UAVs is addressed. Afterward, we detail how to execute the trajectories and control the gimbal. Last, we include a thorough discussion about some critical aspects of the method.

\subsection{Trajectory planning}
\label{sec:trajPlanning}

We plan optimal trajectories for a team of $n$ UAVs as they film a moving actor or target whose position can be measured and predicted. The main objective is to come up with trajectories that satisfy physical UAV and gimbal restrictions, avoid collisions and respect cinematographic concepts. This means that each UAV needs to perform the kind of motion imposed by its shot type (e.g., stay beside/behind the target in a \emph{lateral/chase} shot) and generate smooth trajectories to minimize jerky movements of the camera and yield a pleasant video footage.  
Each UAV will have a shot type and a desired 3D position ($\mathbf{p}_D$) and velocity ($\mathbf{v}_D$) to be reached. This desired state is determined by the type of shot and may move along with the receding horizon. For instance, in a \emph{lateral} shot, the desired position ($\mathbf{p}_D$) moves with the target, to place the UAV beside it; whereas in a \emph{flyby} shot, this position is such that the UAV gets over the target by the end of the shot. More details about the different types of shot and how to compute the desired position will be given in Section~\ref{sec:system_architecture}.  

We plan trajectories for each UAV in a distributed manner, assuming that the plans from other neighboring UAVs are communicated (we denote this set of neighboring UAVs as $Neigh$). For that, we solve a constrained optimization problem for each UAV where the optimization variables are its discrete state with 3D position and velocity ($\mathbf{x}_k = [\mathbf{p}_{Q,k}~\mathbf{v}_{Q,k}]^T$), and its 3D acceleration as control input ($\mathbf{u}_k = \mathbf{a}_{Q,k}$). A non-linear cost function is minimized for a horizon of $N$ timesteps, using as input at each solving iteration the current observation of the system state $\mathbf{x}'$. In particular, the following non-convex optimization problem is formulated for each UAV:

%The discrete state of the system at time step $k$ consists of discrete samples of the UAV position, UAV velocity and target position, respectively: $\mathbf{x}_k = [(\mathbf{p}_{Q,k})^T~(\mathbf{v}_{Q,k})^T~( \mathbf{p}_{T,k})^T]^T$. The dynamics of the system $\mathbf{x}_{k+1} = f(\mathbf{x}_{k}, \mathbf{u}_{k}, \mathbf{v}_T)$, where the control action are $\mathbf{u}_k = \mathbf{a}_{Q,k}$, assume that the target's velocity stays constant $\dot{\mathbf{v}}_T  = 0$ and are obtained by discretizing this equation and \eqref{eq:dynamics} with sampling time $\Delta t$. We use the Runge-Kutta method for integration.  

%%%%%%%%%%%%%%%%%%%%%%%%%%%%%
%% Discrete model for dynamics
%\begin{equation}
%\mathbf{x}_{k+1} = \mathbf{A}\mathbf{x}_{k} + \mathbf{B}\mathbf{u}_{k} 
%\label{eq:model}
%\end{equation}
%\noindent where $\mathbf{u}_k \in \mathbb{R}^3$ are the drone Cartesian accelerations, $\mathbf{A} = 
%\begin{bmatrix}
%\mathbf{I}_3 &  \Delta t \mathbf{I}_3 \\
%\mathbf{0}_3 &  \mathbf{I}_3    \\
%\end{bmatrix}$
% and $\mathbf{B} =
%\begin{bmatrix}
%\frac{\Delta t^2}{2}\mathbf{I}_3 \\
%\Delta t \mathbf{I}_3 \\
%\end{bmatrix}$. 
%%%%%%%%%%%%%%%%%%%%%%%%%%%%%%%%%%%%%%%

\begin{align} 
\label{eq:formulation}
\underset{\begin{subarray}{c}
  \mathbf{x_0},\dots,\mathbf{x_N} \\
  \mathbf{u_0},\dots,\mathbf{u_N}
  \end{subarray}}{\text{minimize}} \ &  \sum_{k=0}^{N} (w_1 ||\mathbf{u}_{k}||^2 + w_2 J_{\theta} + w_3 J_{\psi}) + w_4 J_N& &  \\ 
\text{subject to} \ &\mathbf{x}_0 = \mathbf{x}' & &\text{(\ref{eq:formulation}.a)} \nonumber \\ 
&\mathbf{x}_{k+1} = f(\mathbf{x}_{k},\mathbf{u}_{k}) \quad k = 0,\dots,N-1                        & &\text{(\ref{eq:formulation}.b)} \nonumber \\
&\mathbf{v}_{min} \leq \mathbf{v}_{Q,k} \leq \mathbf{v}_{max} & &\text{(\ref{eq:formulation}.c)} \nonumber \\
&\mathbf{u}_{min} \leq \mathbf{u}_{k} \leq \mathbf{u}_{max} & &\text{(\ref{eq:formulation}.d)} \nonumber \\
&\mathbf{p}_{Q,k} \in \mathcal{F}                               & &\text{(\ref{eq:formulation}.e)} \nonumber \\
&||\mathbf{p}_{Q,k}-\mathbf{p}_{O,k}||^2 \geq r_{col}^2, \quad \forall O  & &\text{(\ref{eq:formulation}.f)} \nonumber \\
&\theta_{min} \leq ^Q\theta_{C,k} \leq \theta_{max} & &\text{(\ref{eq:formulation}.g)} \nonumber \\
&\psi_{min} \leq ^Q\psi_{C,k} \leq \psi_{max} & &\text{(\ref{eq:formulation}.h)} \nonumber \\
&\cos(\beta^{j}_k) \leq cos(\alpha), \quad \forall j \in Neigh & & \text{(\ref{eq:formulation}.i)} \nonumber
\end{align}

As constraints, we impose the initial UAV state (\ref{eq:formulation}.a) and the UAV dynamics (\ref{eq:formulation}.b), which are obtained by integrating numerically the continuous model in Section~\ref{sec:model} with the Runge-Kutta method. We also include  bounds on the UAV velocity (\ref{eq:formulation}.c) and acceleration (\ref{eq:formulation}.d), to ensure trajectory feasibility. 
The UAV position is restricted in two manners. On the one hand, it must stay within the volume $\mathcal{F} \in \mathbb{R}^3$ (\ref{eq:formulation}.e), which is a space not necessarily convex excluding predefined no-fly zones. These are static zones provided by the director before the mission to keep the UAVs away from known hazards like buildings, high trees, crowds, etc. On the other hand, the UAV must stay at a minimum distance $r_{col}$ from any additional obstacle $O$ detected during flight (\ref{eq:formulation}.f), in order to avoid collisions. $\mathbf{p}_{O,k}$ represents the obstacle position at timestep $k$. One of these constraints is added for each other UAV in the team to model them as dynamic obstacles, using their communicated trajectories to extract their positions along the planning horizon. However, other dynamic obstacles, e.g. the actor to be filmed, can also be considered. For that, a model to predict the future position of the obstacle within the time horizon is required. Besides, mechanical limitations of the gimbal to rotate around each axis are enforced by means of bounds on the pitch (\ref{eq:formulation}.g) and yaw angles (\ref{eq:formulation}.h) of the camera with respect to the UAV. Last, there are mutual visibility constraints (\ref{eq:formulation}.i) for each other UAV in the team, to ensure that they do not get into the field of view of the camera at hand. More details about how to compute this constraint are given in Section~\ref{sec:multiUAV}. 

Regarding the cost function, it consists of four weighted terms to be minimized. The terminal cost $J_N=||\mathbf{x}_N-[ \mathbf{p}_D~\mathbf{v}_D]^T ||^2$ is added to guide the UAV to the desired state imposed by the shot type. The other three terms are related with the smoothness of the trajectory, penalizing UAV accelerations and jerky movements of the camera. Specifically, the terms $J_\theta = |^Q\dot{\theta}_{C,k}|^2$ and $J_\psi = |^Q\dot{\psi}_{C,k}|^2$ minimize the angular velocities to penalize quick changes in gimbal angles. Deriving analytically \eqref{eq:Qlambda_C}, $J_\theta$ and $J_\psi$ can be expressed in terms of the optimization variables and the target trajectory. We assume that the target position at the initial timestep is measurable and we apply a kinematic model to predict its trajectory for the time horizon $N$. An appropriate tuning of the different weights of the terms in the cost function is key to enforce shot definition but generating a smooth camera motion.

\subsection{Multi-UAV coordination}
\label{sec:multiUAV}

%% objective
Our method plans trajectories for multiple UAVs as they perform cinematography shots. The cooperation of several UAVs can be used to execute different types of shot simultaneously or to provide alternative views of the same subject. This is particularly appealing for outdoor filming, e.g. in sport events, where the director may want to orchestrate the views from multiple cameras in order to show surroundings during the line of action. In this section, we provide further insight into how we coordinate the motion of the several UAVs while filming. 

%% related work
The first point to highlight is that we solve our optimization problem (\ref{eq:formulation}) on board each UAV in a distributed manner, but being aware of constraints imposed by neighboring teammates. This is reflected in (\ref{eq:formulation}.f) and (\ref{eq:formulation}.i), where we force UAV trajectories to establish a safety distance with others and to stay out of others' field of view for aesthetic purposes. For that, we assume that UAVs are operating close to film the same scene, what allows them to communicate their computed trajectories after each planning iteration. However, there are different alternatives to synchronize the distributed optimization process so that UAVs act in a coordinate fashion. Let us discuss other approaches from key related works and then our proposal. 

In the literature there are multiple works for multi-UAV optimal trajectory planning, but as we showed in Section~\ref{sec:soa}, only few works addressed cinematography aspects specifically. 
A master-slave approach is applied in~\cite{galvane_tg18} to solve conflicts between multiple UAVs. Only one of the UAVs (the master) is supposed to be shooting the scene at a time, whereas the others act as relay slaves that provide complementary viewpoints when selected. The slave UAVs fly in formation with the master avoiding visibility issues by staying out of its field of view. Conversely, fully distributed planning is performed in~\cite{naegeli_tg17} by means of a sequential consensus approach. Each UAV receives the current planned trajectories from all others, and computes a new collision-free trajectory taking into account the whole set of future positions from teammates and the rest of restrictions.  Besides, it is ensured that trajectories for each UAV are planned sequentially and communicated after each planning iteration. In the first iteration, this is equivalent to priority planning, but not in subsequent iterations, yielding more cooperative trajectories.

We follow a hierarchical approach in between. Contrary to~\cite{galvane_tg18}, all UAVs can film the scene simultaneously with no preferences; but there is a scheme of priorities to solve multi-UAV conflicts, as in~\cite{naegeli_tg17}. Thus, the UAV with top priority plans its trajectory ignoring others; the second UAV generates an optimal trajectory applying collision avoidance and mutual visibility constraints given the planned trajectory from the first UAV; the third UAV avoids the two previous ones; and so on.  
This scheme helps coordinating UAVs without deadlocks and reduces computational cost as UAV priority increases. Moreover, we do not recompute and communicate trajectories after each control timestep as in~\cite{naegeli_tg17}; but instead, replanning is performed at a lower frequency and, meanwhile, UAVs execute their previous trajectories as we will describe in next section. 

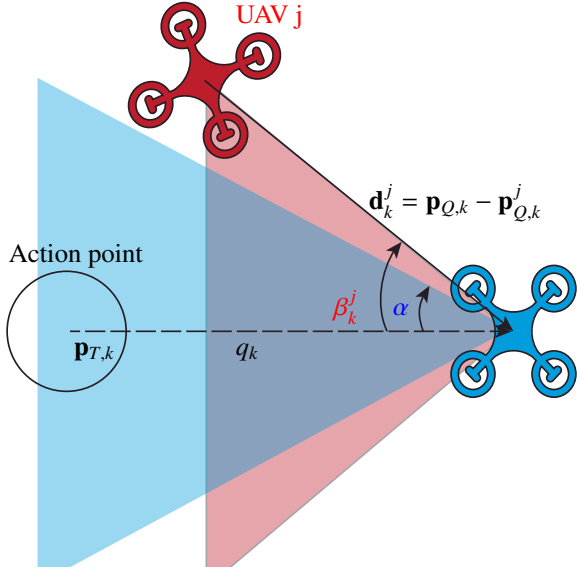
\begin{figure}[tb]
    \centering
    \includegraphics[width=1\columnwidth]{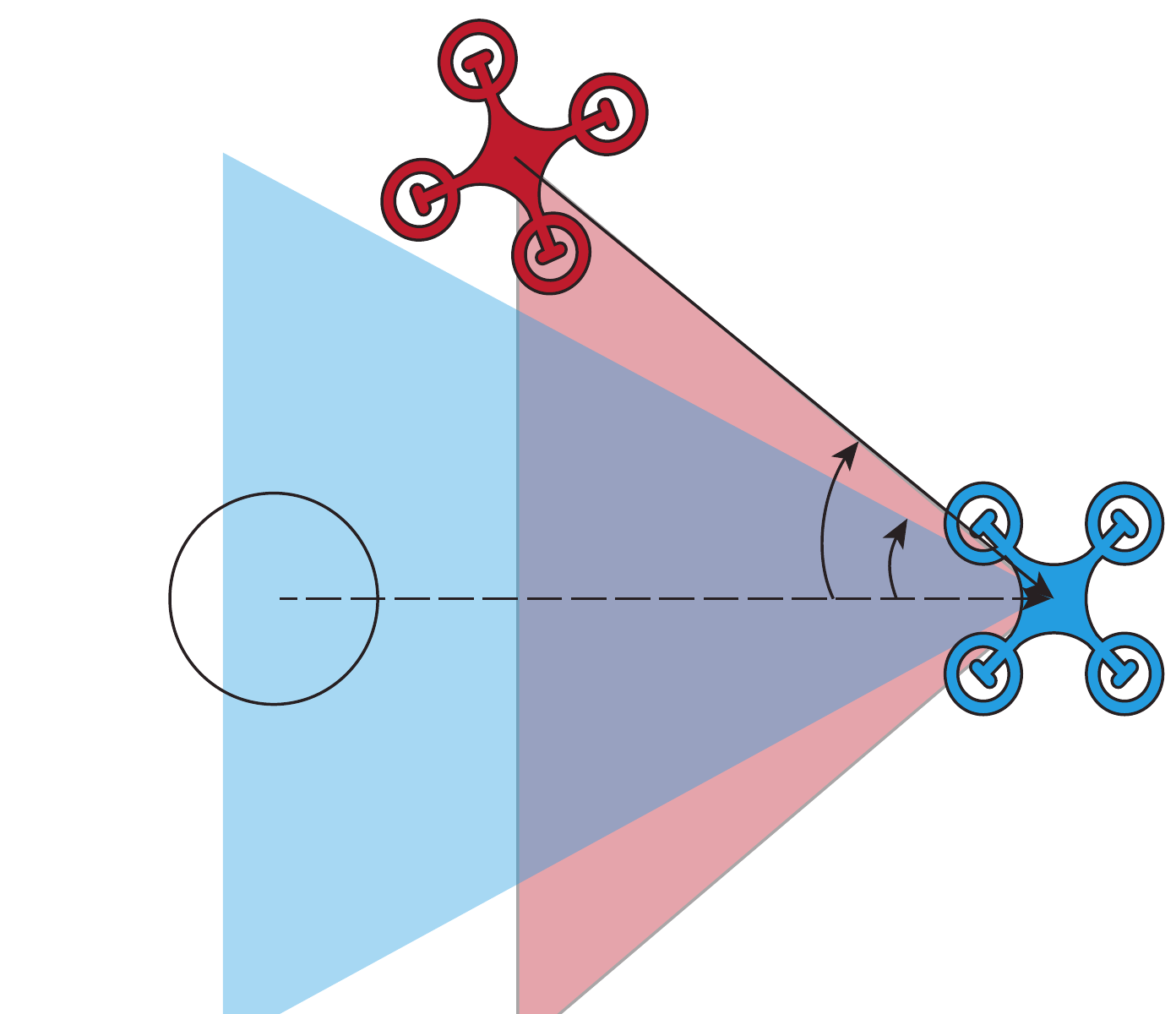}
    \put(-93,95){\textbf{\textcolor{red}{$\beta^{j}_k$}}}
    \put(-71,95){\textbf{\textcolor{blue}{$\alpha$}}}
    \put(-80,135){$\mathbf{d}^{j}_k = \mathbf{p}_{Q,k} - \mathbf{p}_{Q,k}^j$}
    \put(-130,80){$q_k$}
    \put(-193,80){ $\mathbf{p}_{T,k}$}
    \put(-215,115){Action point}
    \put(-130,205){\textcolor{red}{UAV j}}
    \caption{Mutual visibility constraint for two UAVs. The UAV on the right (blue) is filming an action point at the same time that it keeps the UAV on top (red) out of its angle of view $\alpha$.}
    \label{fig:mutual}
\end{figure}

In terms of multi-UAV coordination, constraint (\ref{eq:formulation}.f) copes with collisions between teammates and (\ref{eq:formulation}.i) with mutual visibility. We consider all neighboring UAVs as dynamic obstacles whose trajectories are known (plans are communicated), and we enforce a safety inter-UAV distance $r_{col}$ along the entire planning horizon $N$. The procedure to formulate the mutual visibility constraint is illustrated in Figure~\ref{fig:mutual}. The objective is to ensure that each UAV's camera has not other UAVs within its field of view (the angle of view is denoted as $\alpha$). We approximate the actual field of view of the camera with a circular shape, and $\alpha$ is the semi-cone angle of the cone surrounding the real field of view. We think this is a good approximation for long-range shots and it simplifies the formulation of the mutual visibility constraints, which alleviates the problem non-linearity and helps computing a solution. Geometrically, we model UAVs as points that need to stay out of the field of view, but select $\alpha$ large enough to account for UAV dimensions. If we consider the UAV that is planning its trajectory at position $\mathbf{p}_{Q,k}$ and another neighboring UAV $j$ at position $\mathbf{p}_{Q,k}^j$, then $\beta^{j}_k$ refers to the angle between vectors 
$\mathbf{q}_k = \mathbf{p}_{Q,k} - \mathbf{p}_{T,k}$ and $\mathbf{d}^{j}_k = \mathbf{p}_{Q,k} - \mathbf{p}_{Q,k}^j$:

\begin{equation}
\cos(\beta^{j}_k) = \frac{\mathbf{q}_k \cdot \mathbf{d}_k^{j}}{|| \mathbf{q}_k|| \cdot || \mathbf{d}_k^{j} || },
\end{equation}

\noindent being $\cos (\beta_k^j) \leq \cos (\alpha)$ the condition to keep UAV $j$ out of the field of view. 

Finally, it is important to notice that there may be certain situations where our priority scheme to apply mutual visibility constraints could fail. If we plan a trajectory for the UAV with priority 1, and then, another one for the UAV with lower priority 2; ensuring that UAV 1 is not within the field of view of UAV 2 does not imply the way around, i.e., UAV 2 could still appear on UAV 1's video. However, these situations are rare in our cinematography application, as there are not many cameras pointing in random directions, but only a few and all of them filming a target typically on the ground. Moreover, since we favor smooth trajectories, we experienced in our tests that our solver tends to avoid crossings between different UAVs' trajectories, as that would result in more curves. Therefore, establishing UAV priorities in a smart way, based on their height or distance to the target, was enough to prevent issues related with mutual visibility.

\subsection{Trajectory execution}
\label{sec:trajExecution}

Our trajectory planners produce optimal trajectories containing UAV positions and velocities sampled at the control timestep, which we can denote as $\Delta t$. As we do not recompute trajectories at each control timestep for computational reasons, we use another independent module for trajectory following, whose task is flying the UAV along its current planned trajectory. This module is executed at a rate of $1/\Delta t~Hz$ and keeps a track of the last computed trajectory, which is replaced after each planning iteration. Each trajectory follower computes 3D velocity references for the velocity controller on board the UAV. For this purpose, we take the closest point in the trajectory to the current UAV position, and then, we select another point in the trajectory at least $L$ meters ahead. The 3D velocity reference is a vector pointing to that \emph{look-ahead} waypoint and with the required speed to reach the point within the specified time in the planned trajectory. 

At the same time that UAVs are following their trajectories, a gimbal controller is executed at a rate of $1/\Delta t_G$ $Hz$ to point the camera toward the target being filmed. 
We assume that the gimbal has an IMU and a low-level controller receiving angular rate commands, defined with respect to the world reference frame $\lbrace W \rbrace$. Using feedback about the target position, we generate references for the gimbal angles to track the target and compensate the UAV motion and possible errors in trajectory planning. These references are sent to an attitude controller that computes angular velocity commands based on the error  between current and desired orientation in the form of a rotation matrix $R_e = (R_C)^T R_C^*$, where the desired rotation matrix $R_C^*$ is given by \eqref{eq:RC}. Recall that we assumed that $R_C$ instantaneously takes the value of $R_C^*$. To design the angular velocity controller, we use a standard first-order controller for stabilization on the Special Orthogonal Group $\mathbb{SO}(3)$, which is given by $\boldsymbol{\omega} = k_\omega (R_e-R_e^T)^\vee$, where the \emph{vee} operator $\vee$ transforms $3\times3$ skew-symmetric matrices into vectors in $\mathbb{R}^3$ \cite{Lee2013}.
More specific details about the mathematical formulation of the gimbal controller can be seen in~\cite{cunha_eusipco19}.

%%%%%%%%%%%%%%%%%%%%%%%%%%%%%%%%%%%%%%%%%%%%%%%%
%%%%%%%%%%%%%%%%%%%%%%%%%%%%%%%%%%%%%%%%%%%%%%%%

\subsection{Discussion}
\label{sec:discussion}

In this section, we discuss some critical aspects of our method for trajectory planning. In particular, its optimality and convergence time, as well as how it deals with issues such as delays computing solutions, external disturbances due to bad weather or obstacle representation.

\paragraph{Optimality} We apply numerical methods to solve the optimization problem described in Section~\ref{sec:trajPlanning}, thus converging to an optimal solution for a single UAV. Even though there are no theoretical guarantees of achieving the global optimum when solving a non-linear and non-convex optimization problem, we experienced good results with the numerical solver that we used both in terms of local optimality and computation time. A proper solver initialization is essential for fast convergence, so we use the last computed trajectory to initialize the solution search. Nonetheless, as we are considering a formulation with multiple UAVs acting simultaneously, our method does not achieve the optimal solution for the complete team. This is because we impose a priority scheme and solve each UAV trajectory assuming others' trajectories fixed for the given time horizon. Even though it would be more optimal to recompute and exchange solutions after each execution time step for all UAVs~\cite{naegeli_tg17}, the quality of our solutions was enough for the purpose of the application. Moreover, in our experiments, UAV priorities were fixed, but the method could be adapted easily to consider priorities that vary during the mission depending on certain circumstances to be more efficient. We leave as future work a further analysis to establish bounds on the quality degradation of our solution compared with the complete multi-UAV optimum.

\paragraph{Convergence time} Our trajectory planning problem is a non-linear and non-convex optimization that is complex to solve; even if the team of UAVs does not encounter external obstacles, they need to consider inter-UAV collision avoidance and mutual visibility. Therefore, the time to converge to a solution is not negligible. We tackle this by limiting the time horizon for trajectory planning (which reduces computation time) and using different rates for trajectory planning and execution. Trajectory planning is performed at lower rates to reduce computation (between $0.5$ and $2Hz$ in our experiments). Besides, we limit the computation time for the solver and keep following the last computed trajectory until it converges to a new solution. In case that the maximum computation time is reached without convergence, there are no guarantees regarding the quality of the solution computed, so we recalculate changing the problem initialization with the current UAV state, which is usually enough to converge to a new solution. In the unlikely case of reaching the end of the previous computed trajectory without new solution, the UAV would stay hovering and recomputing trajectories with different initial solutions until convergence.\\ In addition, we do not assume that solutions are generated instantly and we deal with delays when planning trajectories. The generated trajectories have time stamps associated with each waypoint. The trajectory follower component described in Section~\ref{sec:trajExecution} receives these trajectories with certain delay (due to the solution computation time) and synchronizes them by discarding the initial waypoints corresponding to time instants already gone by.

\paragraph{Performance under external perturbations} Keeping flight stability and smooth trajectories even under external disturbances such as bad weather conditions is critical in our method.
In the presence of bad weather, the trajectory planning components (Section~\ref{sec:trajPlanning}) would still generate smooth trajectories; however, windy conditions could result in an inaccurate trajectory following due to external perturbations. Therefore, the key to improve stability under bad weather conditions would be implementing more robust UAV controllers. In our case, we implemented a trajectory follower based on a \emph{pure pursuit} algorithm with a \emph{look-ahead} parameter and a velocity controller. Nonetheless, alternative control techniques~\cite{kamel_iros17,kostadinov_iros20} taking into account external perturbations and uncertainties or integrating non-linear models for the UAV could be applied to increase flight stability in case of wind gusts. 
Besides, in terms of trajectory planning, we could also adapt the weights of the cost function in case of bad weather, penalizing more those costs based on UAV  accelerations and gimbal angular velocities, and relaxing the cost associated with the desired final state. Thus, the generated trajectories would be more conservative from the smoothness point of view, which would help following trajectories in these adverse conditions.

\paragraph{Obstacle representation} In our problem formulation, we include predefined no-fly zones and additional dynamic obstacles. The former are used to indicate static hazards with known positions, like buildings, areas with trees, etc. The latter consists of other UAVs in the team or external obstacles, e.g., the target being filmed, other actors in the scene, etc. As explained in Section~\ref{sec:trajPlanning}, we represent these dynamic obstacles by means of spherical objects of radius $r_{col}$, since the constraint included is to keep that safety distance between the 3D obstacle position and the corresponding UAV. We also explained that we need a prediction model to estimate object trajectories within the planning horizon time. We use a constant velocity model to compute those future predictions, although more complex models could be used too.
Moreover, alternative geometrical representations could be used for the obstacles if more information about their shape were known. For instance, 3D ellipsoids with three different axis lengths are used in \cite{naegeli_tg17}. In our context, we do not foresee UAVs getting so close to targets so that its geometrical shape really matters, and hence, we preferred spherical shapes that ease mathematical formulation.

Obstacle detection is out of the scope of this paper, so we assume that there is a perception module providing an estimation of the obstacle 3D positions and velocities (for motion prediction). In practice, we used in our experiments dynamic obstacles whose positions could be measured with a GPS and communicated, i.e., other UAV teammates and the filmed target. Nonetheless, this information could be obtained by algorithms processing measurements from pointcloud-based sensors on board the UAVs, such as 3D LIDARs or RGB-D cameras. In that case, alternative obstacle representations based on distance to the points (e.g., to the centroid or to the closest point) within the corresponding pointclouds could be used, as it is done by the authors in \cite{kratky_ral20}.

\section{System Architecture}
\label{sec:system_architecture}

\begin{figure*}[tb!]
    \centering
    \includegraphics[width=2\columnwidth]{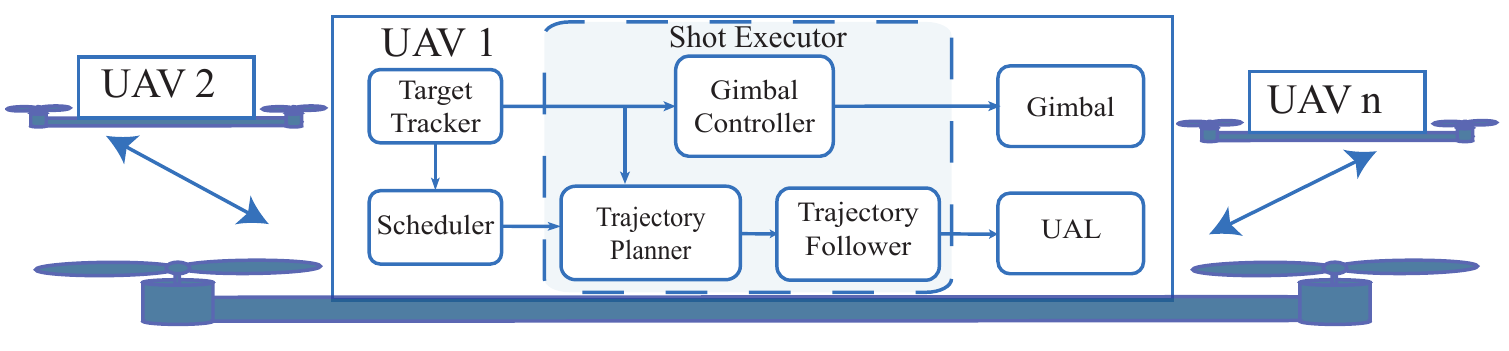}
    \put(-333,55){\small $\mathbf{p}_D$}
    \put(-333,45){\small $\mathbf{v}_D$}
    \put(-333,94){\small $\mathbf{p}_T$}
    \put(-333,84){\small $\mathbf{v}_T$}
    % \put(-327,75){\small $\mathbf{v}_T$}
    % \put(-186,35){\small $\mathbf{v}_C$}
    %\put(-185,75){\small Gimbal Commands}
    
    \caption{System architecture on board each UAV. A Scheduler initiates the shot and updates continuously the desired state for trajectory planning, whereas the Shot Executor plans optimal trajectories to perform the shot. UAVs exchange their plans for coordination.}
    \label{fig:block_diagram}
\end{figure*}
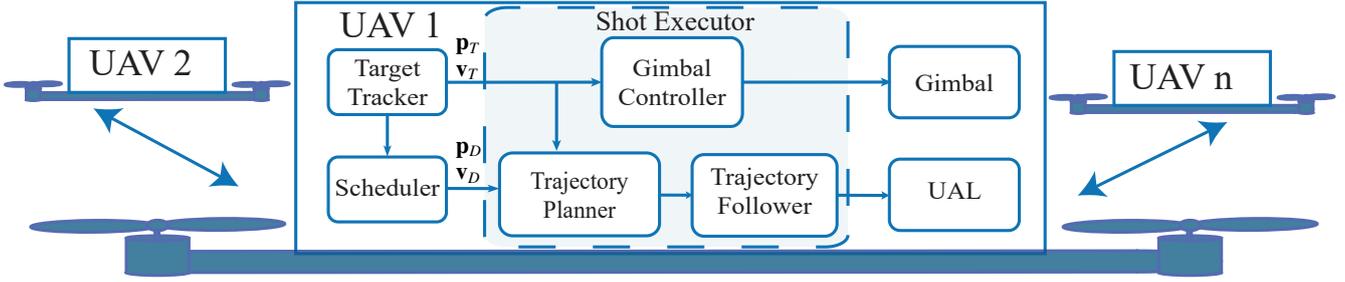

In this section, we present our system architecture, describing the different software components required for trajectory planning and their interconnection. Besides, we introduce briefly the overall architecture of our complete system for cinematography with multiple UAVs, which was presented in~\cite{alcantara_access20}.

Our system counts on a \textit{Ground Station} where the components related with mission design and planning are executed. We assume that there is a cinematography \textit{director} who is in charge of describing the desired shots from a high-level perspective. We created a graphical tool and a novel cinematography language~\cite{montes_appsci20} to support the director through this task. Once the mission is specified, the system has planning components~\cite{caraballo_iros20} that compute feasible plans for the mission, assigning shots to the available UAVs according to shot duration and remaining UAV flight time. The mission execution is also monitored in the Ground Station, in order to calculate new plans in case of unexpected events like UAV failures.  

The components dedicated to shot execution run on board each UAV. Those components are depicted in Figure~\ref{fig:block_diagram}. Each UAV has a  \textit{Scheduler} module that receives shot assignments from the Ground Station and indicates when a new shot should be started. Then, the \textit{Shot Executor} is in charge of planning and executing optimal trajectories to perform each shot, implementing the method described in Section~\ref{sec:method}. As input, the Shot Executor receives the future desired 3D position $\mathbf{p}_D$ and velocity $\mathbf{v}_D$ for the UAV, which is updated continuously by the Scheduler 
depending on the shot parameters and the target position. For instance, in a lateral shot, the dynamic model of the target is used to predict its position by the end of the horizon time and then place the UAV desired position at the lateral distance indicated by the shot parameters. 

Additionally, the target positioning provided by the \textit{Target Tracker} is required by the Shot Executor to point the gimbal and place the UAV adequately. In order to alleviate the effect of noisy measurements when controlling the gimbal and to provide target estimations at high frequency, the Target Tracker implements a Kalman Filter integrating all received observations. This filter is able to accept two kinds of measurements: 3D global positions coming from a GPS receiver on board the target, and 2D positions on the image obtained by a vision-based detector~\cite{nousi_icip19}. 
In particular, in the experimental setup for this paper, we used a GPS receiver on board a human target communicating measurements to the Target Tracker. Communication latency and lower GPS rates are addressed by the Kalman Filter to provide a reliable target estimation at high rate. 

The Shot Executor, as it was explained in Section~\ref{sec:method}, consists of three submodules: the \emph{Trajectory Planner}, the \emph{Trajectory Follower} and the \emph{Gimbal Controller}. The Trajectory Planner computes optimal trajectories for the UAV solving the problem in \eqref{eq:formulation} in a receding fashion, trying to reach the desired state indicated by the Scheduler. The Trajectory Follower calculates 3D velocity commands at higher rate so that the UAV follows the optimal reference trajectory, which is updated any time the Planner generates a new solution. The Gimbal Controller generates commands for the gimbal motors in the form of angular rates in order to keep the camera pointing towards the target. 
The \emph{UAV Abstraction Layer} (UAL) is a software component developed by our lab~\cite{real_ijars20} to interface with the position and velocity controllers of the UAV autopilot. It provides a common interface abstracting the user from the protocol of each specific hardware. Finally, recall that each UAV has a communication link with other teammates in order to share their current computed trajectories, which are used for multi-UAV coordination by the Trajectory Planner.

%%%%%%%%%%%%%%%%%%%%%%%%%%%%%%%%%%%%%%%%%%%%%%%%%
%%%%%%%%%%%%%%%%%%%%%%%%%%%%%%%%%%%%%%%%%%%%%%%%%

\subsection{Cinematography shots}
\label{sec:shots}

In our previous work~\cite{alcantara_access20}, following recommendations from cinematography experts, we selected a series of canonical shot types for our autonomous multi-UAV system. Each shot has a type, a time duration and a set of geometric parameters that are used by the system to compute the desired camera position with respect to the target. 
The representative shots used in this work for evaluation are the following:

\begin{itemize}
    \item \textit{Chase/lead:} The UAV chases a target from behind or leads it in the front at a certain distance and with a constant altitude.
    
    \item \textit{Lateral:} The UAV flies beside a target with constant distance and altitude as the camera tracks it. 
    
    \item \textit{Flyby:} The UAV flies overtaking a target with a constant altitude as the camera tracks it. The initial distance behind the target and final distance in front of it are also shot parameters. 
    
    \item \textit{Orbit:} The UAV flies with a constant altitude orbiting around the target from a certain distance, as the camera tracks it.
 
\end{itemize}

Even though our complete system~\cite{alcantara_access20} implements additional shots, such as static, elevator, etc., they follow similar behaviors or are not relevant for trajectory planning evaluation. Particularly, we distinguish between two groups of shots for assessing the performance of the trajectory planner: (i) shots where the relative distance between UAV and target is constant (e.g., chase, lead or lateral), denoted as \emph{Type I} shots; and (ii) shots where this relative distance varies throughout the shot (e.g., flyby or orbit), denoted as \emph{Type II} shots. Note that an orbit shot can be built with two consecutive flyby shots. In Type I shots, the relative motion of the gimbal with respect to the UAV is quite limited, and the desired camera position does not vary with the shot phase, i.e., it is always at the same distance of the target. In Type II shots though, there is a significant relative motion of the gimbal with respect to the UAV, and the desired camera position depends on the shot phase, e.g., it transitions from behind to the front throughout a flyby shot. These two kinds of patterns will result in different behaviors of our trajectory planner, so for a proper evaluation, we test it with shots from both groups. 

%%%%%%%%%%%%%%%%%%%%%%%%%%%%%%%%%%%%%%%%%%%%%
%%%%%%%%%%%%%%%%%%%%%%%%%%%%%%%%%%%%%%%%%%%%%

\section{Performance Evaluation}
\label{sec:simulations}

%%%%%%%%%%%%%%%%%%% simulations to perform %%%%%%%%%%%%%%%%%%%%%%%%%%%%%%%%%%%%%

%%% In \cite{bonnati_jfr20} they evaluate quality of trajectories with normalized cost: J/horizon. See table 10. They estimate that horizons of around 10 seconds are needed for cinematography. They run a user study in Section 9.4.2 to evaluate their algorithm to select shots. They include people comments.
%% we could compare our trajectories with normal executer from IST. Reproduce our typical trajectories in a nice simulator to generate videos. Test if jerk is different?

%%%%%%%%%%%%%%%%%%%%%%%%%%%%%%%%%%%%%%%%%%%%%%%%%%%%%%%%%%%%%%%%%%%%%%%%%%%%%

In this section, we present experimental results to assess the performance of our method for trajectory planning in cinematography. We evaluate the behavior of the resulting trajectories for the two categories of shots defined, demonstrating that our method achieves smooth and less jerky movements for the cameras. We also show the effect of considering physical limits for gimbal motion, as well as multi-UAV constraints due to collision avoidance and mutual visibility.  

We implemented our trajectory planner described in Section~\ref{sec:method} by means of \emph{Forces Pro}~\cite{FORCESnlp}, which is a software that creates domain-specific solvers in C language for non-linear optimization problems. Forces Pro uses direct multiple shooting~\cite{multiple_shooting} for problem discretization, approximating the state trajectories to achieve a finite-dimensional optimization problem. Then, an algorithm based on the interior-point method is used to solve this non-linear optimization. Table~\ref{tab:parameters} depicts common values for some parameters of our method used in all the experiments, where physical constraints correspond to our actual UAV prototypes. Moreover, all the experiments in this section were performed with a MATLAB-based simulation environment integrating the C libraries from Forces Pro, in a computer with an Intel Core i7 CPU @ $3.20~GHz$, $8~Gb$ RAM.     

\begin{table}[]
    \centering
    \begin{tabular}{|c|c|}
    \hline
    Parameter & Value \\ \hline \hline
    $\mathbf{u}_{min},\mathbf{u}_{max}$ & $\pm 5~m/s^2$ \\ \hline
    $\mathbf{v}_{min},\mathbf{v}_{max}$  & $\pm 10~m/s$ \\ \hline
    %$r_{col}$ & $4~m$ \\
    $\theta_{min},\theta_{max}$ & $-\pi/2,-\pi/4~rad$ \\ \hline
    $\psi_{min},\psi_{max}$ & $-3\pi/4,3\pi/4~rad$ \\ \hline
    $\alpha$ & $\pi/6~rad$\\ \hline
    $\Delta t$, $\Delta t_G$         & $0.1s$ \\ \hline
    $L$ & $1~m$\\
    \cline{1-2}
    \end{tabular}
    \caption{Common values of method parameters in experiments.}
    \label{tab:parameters}
\end{table}{}

\subsection{Cinematographic aspects}

First, we evaluate the effect of imposing cinematography constraints in UAV trajectories. For that, we selected a shot of Type II, since their relative motion between the target and the camera makes them richer to analyze cinematographic effects.
Particularly, we performed a flyby shot with a single UAV, filming a target that moves on the ground with a constant velocity ($1.5~m/s$) along a straight line (this constant motion model is used to predict target movement).  
The UAV had to take a shot of $10$ seconds at a constant altitude of $3~m$, starting $20~m$ behind the target and overtaking it to end up $15~m$ ahead. Moreover, we placed a circular no-fly zone at the starting position of the target, simulating the existence of a tree.    

\begin{figure*}[]
    \begin{subfigure}[b]{0.475\textwidth}
        \includegraphics[width=1\textwidth]{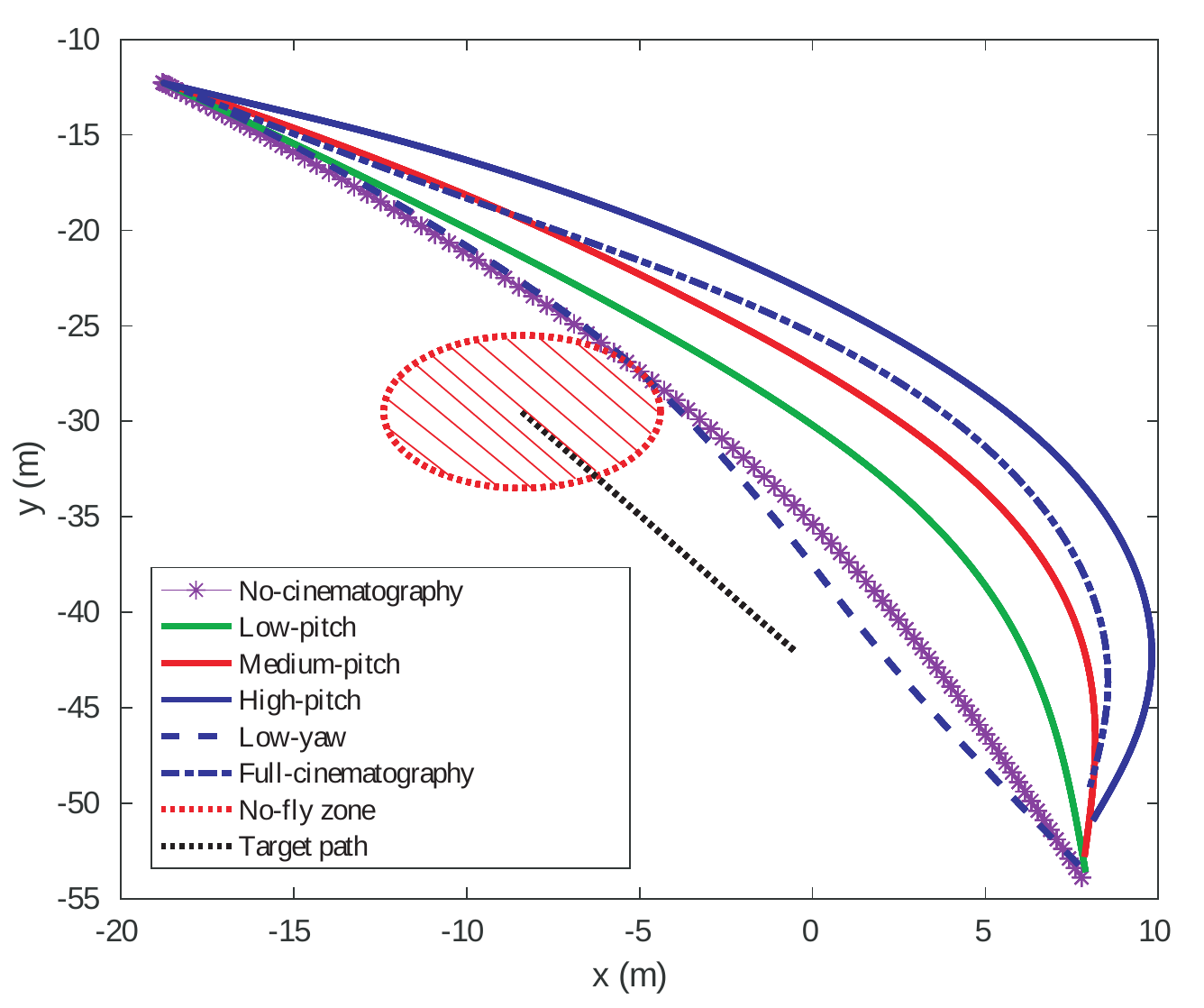}
        \centering
        \put(-55,25){$\mathbf{p}_D$}
        \put(-218,165){$\mathbf{p}_{Q,0}$}
        \put(-100,65){$\mathbf{p}_{T,N}$}
        \put(-160,120){$\mathbf{p}_{T,0}$}
    \end{subfigure}
    \begin{subfigure}[b]{0.505\textwidth}
            \centering
            \includegraphics[width=1\textwidth]{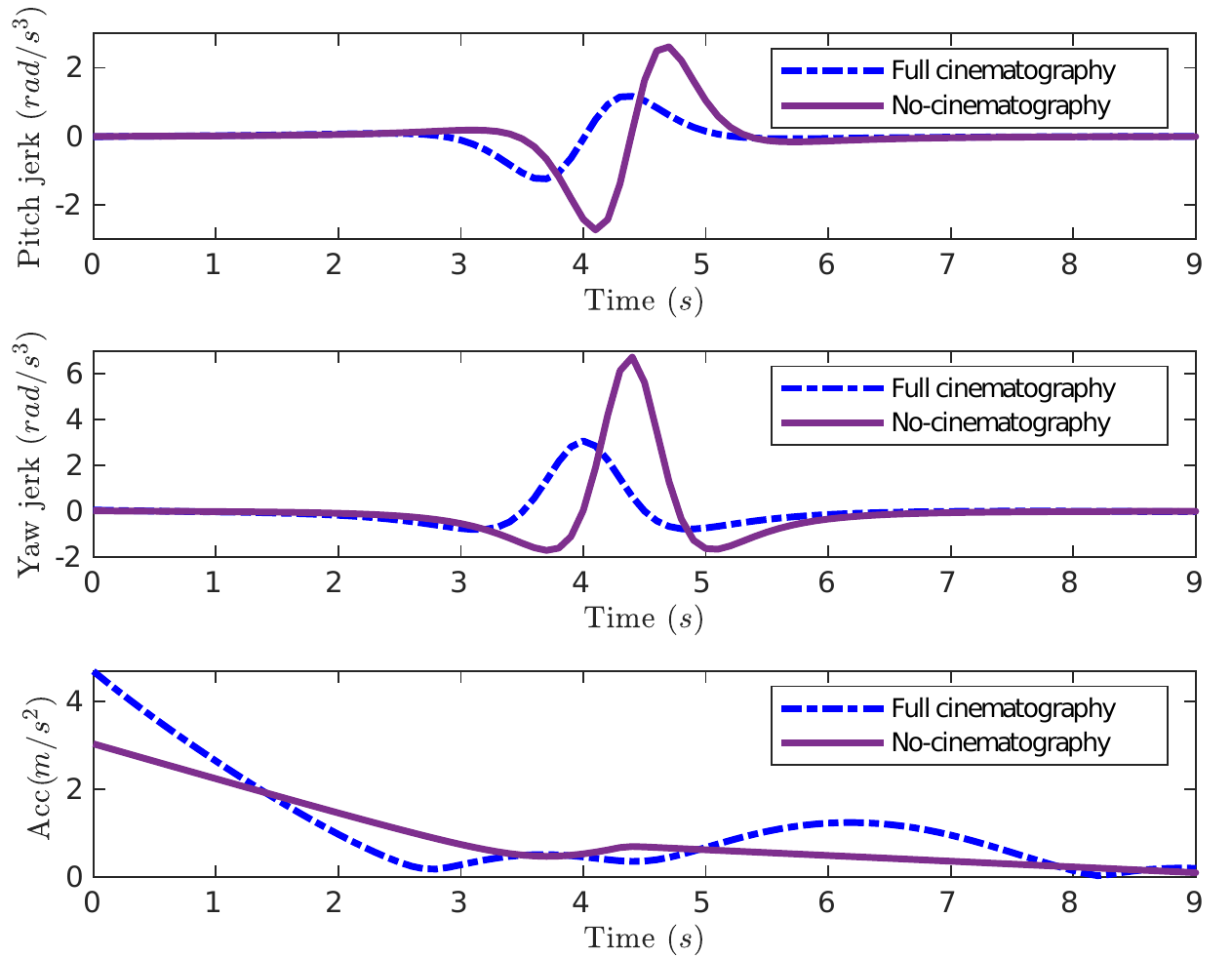}
    \end{subfigure}
    \caption{Left, top view of the resulting trajectories for different solver configurations for the cost weights. High-yaw is not shown as it was too similar to low-yaw. The target follows a straight path and the UAV has to execute a flyby shot ($10~s$) starting $20~m$ behind and ending up $15~m$ ahead. The predefined no-fly zone simulates the existence of a tree. Right, temporal evolution of the jerk of the camera angles and the norm of the 3D acceleration. We compare the \emph{full-cinematography} configuration against \emph{no-cinematography}.}
    \label{fig:simFlyover}
\end{figure*}

We evaluated the quality of the trajectories computed by our method setting the horizon to $N=100$, in order to calculate the whole trajectory for the shot duration ($10~s$) in a single step, instead of using a receding horizon~\footnote{The average time to compute each trajectory was $\sim 100~ms$, which allows for online computation.}. We tested different configurations for comparison: \emph{no-cinematography} uses $w_2=w_3=0$; \emph{low-pitch}, \emph{medium-pitch} and \emph{high-pitch} use $w_3=0$ and $w_2=100$, $w_2=1\,000$ and $w_2=10\,000$, respectively; \emph{low-yaw} and \emph{high-yaw} use $w_2=0$ and $w_3=0.5$ and $w_3=1$, respectively; and \emph{full-cinematography} uses $w_2=10\,000$ and $w_3=0.5$. For all configurations, we set $w_1=w_4=1$. These values were selected empirically to analyze the planner behavior under a wide spectrum of weighting options in the cost function.  
Figure~\ref{fig:simFlyover} (left) shows the trajectory followed by the target and the UAV trajectories generated with the different options. Even though trajectory planning is done in 3D, the altitude did not vary much, as the objective was to perform a shot with a constant altitude. Therefore, a top view is depicted to evaluate better the effect of the weights. 

Table~\ref{tab:metrics} shows a quantitative comparison of the different configurations. For this comparison, we define the following metrics. First, we measure the minimum distance to any obstacle or no-fly zone in order to check collision avoidance constraints.
Then, we measure the average norm of the 3D acceleration along the trajectory, and of the jerk (third derivative) of the camera angles $\theta_C$ and $\psi_C$. These three metrics provide an idea on whether the trajectory is smooth and whether it implies jerky movement for the camera. Note that jerky motion has been identified in the literature on aerial cinematography~\cite{bonatti_jfr20,gebhardt_chi16} as a relevant cause for low video quality. Figure~\ref{fig:simFlyover} (right) depicts the temporal evolution of jerk of the camera angles and the norm of the 3D acceleration.
%We also measure the \emph{Root Mean Square Error} (RMSE) of the generated trajectory with respect to the trajectory that the UAV would follow without considering cinematographic nor collision avoidance aspects. This latter trajectory is included in Table~\ref{tab:metrics} as \emph{baseline}. For example, in the considered flyby shot, this would be the UAV overtaking the target flying on top of it, i.e., a UAV motion along the straight line that the target follows but at a higher altitude. This is the shorter way to execute the flyby, but our solver will produce different trajectories, as that option is not smooth and breaks physical constraints of the gimbal. 

\begin{table}
\renewcommand*{\arraystretch}{1.5}
\centering
\resizebox{\columnwidth}{!}{%
\begin{tabular}{|c|c|c|c|c| } 
 \cline{2-5} 
 \multicolumn{1}{c|}{} &			Dist ($m$) & Acc ($m/s^2$) & Yaw jerk ($rad/s^3$) & Pitch jerk ($rad/s^3$) \\ \cline{2-5} \hline 
 \textit{No-cinematography} & 0.00  & 0.86 & 0.61 & 0.31 \\ \hline
 \textit{Low-pitch} 		     & 1.81  & 1.13 &  0.35 & 0.15   \\ \hline
 \textit{Medium-pitch} 	     &  6.38 & 1.25 & 0.19 & 0.07 \\ \hline
 \textit{High-pitch}	         & 6.13& 1.42& 0.10 & 0.03 \\ \hline
 \textit{Low-yaw}	&      0.00  & 0.81 & 0.52 & 0.25 \\ \hline
 \textit{High-yaw}   & 0.00  &1.00 & 0.50 & 0.23\\ \hline
 \textit{Full-cinematography} & 4.44 & 1.27 & 0.10  & 0.03\\ \hline
 \textit{Full-cinematography (receding)} & 4.19 &1.45 &0.08 &0.03\\ \hline
\end{tabular}}
\caption{Resulting metrics for flyby shot. \emph{Dist} is the minimum distance to the no-fly zone. \emph{Acc}, \emph{Yaw jerk} and \emph{Pitch jerk} are the average norms along the trajectory of the 3D acceleration and the jerk of the camera yaw and pitch, respectively.\label{tab:metrics}}
\end{table}

Our experiment allows us to derive several conclusions. 
The \emph{no-cinematography} configuration produces a trajectory that gets as close as possible to the no-fly zone and minimizes 3D accelerations (curved). However, when increasing the weight on the pitch rate, trajectories get further from the target and accelerations increase slightly (as longer distances need to be covered in the same shot duration), but jerk in camera angles is reduced. On the contrary, activating the weight on the yaw rate, trajectories get closer to the target again. With the \emph{full-cinematography} configuration, we achieve the lowest values in angle jerks and a medium value in 3D acceleration, which seems to be a pretty reasonable trade-off. It can also be seen in Figure~\ref{fig:simFlyover} (right) how this configuration reduces camera acceleration and angle jerks with respect to \emph{no-cinematography}, obtaining smoother trajectories.  

Finally, we also tested the \emph{full-cinematography} configuration in a receding horizon manner. In that case, the solver was run at $1~Hz$ with a time horizon of $5~s$ ($N=50$). The resulting metrics are included in Table~\ref{tab:metrics}. Using a receding horizon with a horizon shorter than the shot's duration is suboptimal, and average acceleration increases slightly. However, we achieve similar values of angle jerk, plus a reduction in the computation time~\footnote{The average time to compute each trajectory was $\sim 7 ms$.}. Moreover, this option of recomputing trajectories online would allow us to correct possible deviations on predictions for the target motion, in case of more random movements (in these simulations, the target moved with a constant velocity).  

\subsection{Time horizon}

\begin{figure}[tb!]
    \centering
    \includegraphics[width=1\columnwidth]{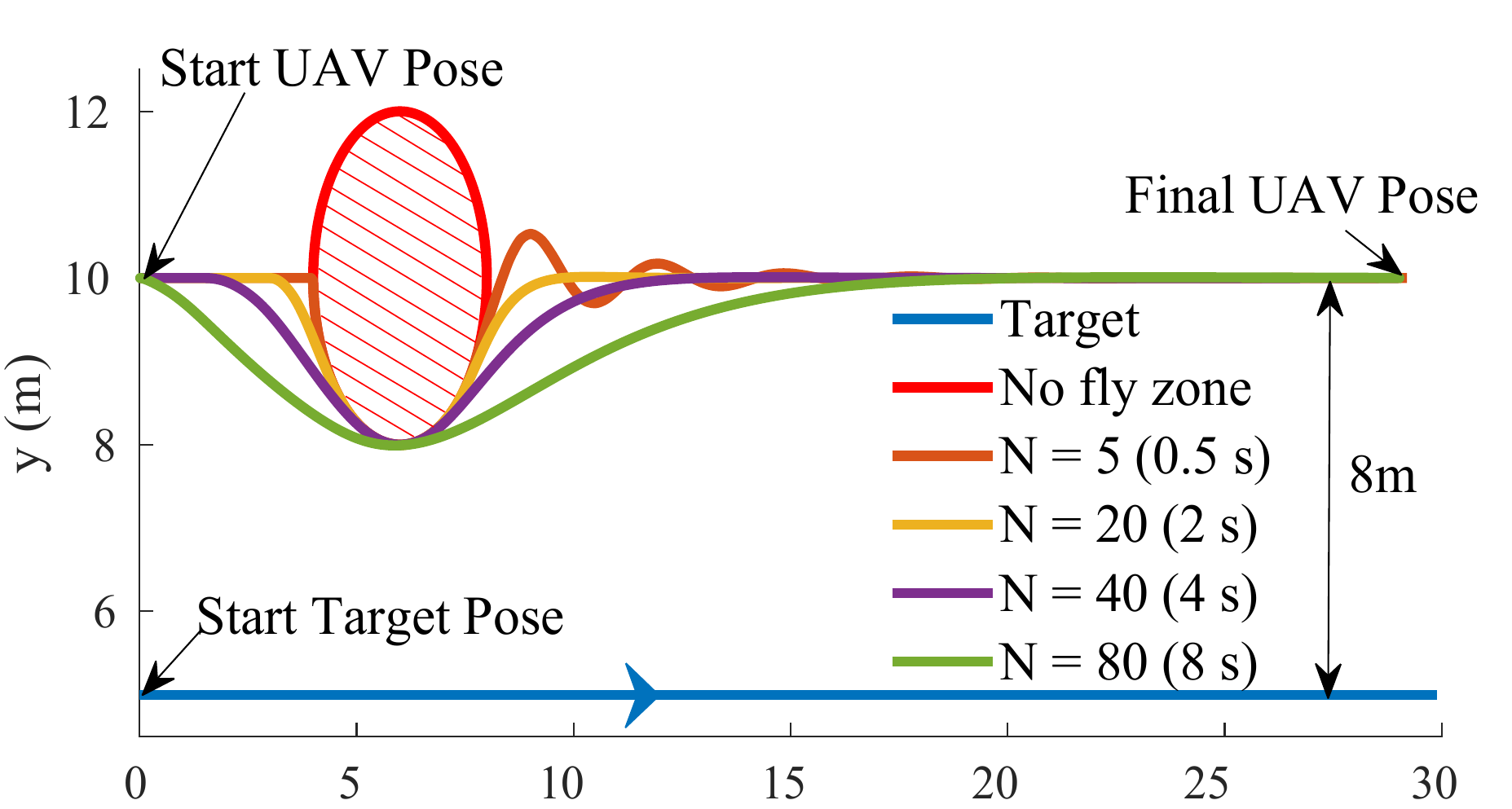}
    \caption{Receding horizon comparative for lateral shot. Top view of the resulting trajectories for different time horizons. The target follows a straight path and the UAV has to execute a lateral shot ($10~s$) at a $8~m$ distance.}
    \label{fig:lateral_shot_simulation}
\end{figure}

We also performed another experiment to evaluate the performance of shots of Type I. In particular, we selected a lateral shot to show results, but the behavior of other shots like chase or lead was similar, as they all do the same but with a different relative position w.r.t. the target. We executed a lateral shot with a duration of $20~s$ to film a target from a lateral distance of $8~m$ and a constant altitude of $3~m$. As in the previous experiment, the target moves on the ground with a constant velocity ($1.5~m/s$) along a straight line, and we used that motion model to predict its movement. In normal circumstances, the type of trajectories followed to film the target are not so interesting, as the planner only needs to track it laterally at a constant distance. Therefore, we used this experiment to analyze the effects of modifying the time horizon, which is a critical parameters in terms of both computational load and capacity of anticipation. We used our solver in receding horizon recomputing trajectories at $2~Hz$, and we placed a static no-fly zone in the middle of the UAV trajectory to check its avoidance during the lateral shot under several values of the time horizon $N$. The top view of the resulting trajectories (the altitude did not vary significantly) can be seen in Figure~\ref{fig:lateral_shot_simulation}. % We used the low-pitch configuration for cost function weights but similar results were obtained with other configurations. 
Table~\ref{tab:metricsLateral} shows also the performance metrics for the different trajectories.  

We can conclude that trajectories with a longer time horizon were able to predict the collision more in advance and react in a smoother manner; while shorter horizons resulted in more reactive behaviors.
In general, for the kind of outdoor shots that we performed in all our experiments, we realized that time horizons in the order of several seconds (in consonance with~\cite{bonatti_jfr20}) were enough, as the dynamics of the scenes are not extremely high. We also tested that the computation time for our solver was short enough to calculate these trajectories online.   

\begin{table}
\centering
\renewcommand*{\arraystretch}{1.5}
\resizebox{\columnwidth}{!}{%
\begin{tabular}{ |c|c|c|c|c| } 
 \cline{1-5} Time horizon ($s$) 
 &   Acc ($m/s^2$) & Yaw jerk ($rad/s^3$) & Pitch jerk ($rad/s^3$) & Solution time ($s$) \\ \hline\hline
 0.5 &  0.15 & 10.68      & 1.35 & 0.004        \\ \hline
 2 & 0.08 & 5.12  & 0.32 & 0.016 \\ \hline
 4 		   & 0.05 &  5.06 & 0.32 & 0.029   \\ \hline
 8 	     & 0.04 & 4.8 & 0.29 & 0.101 \\ \hline
\end{tabular}}
\caption{Resulting metrics for the lateral shot. \emph{Solution time} is the average time to compute each trajectory.
}\label{tab:metricsLateral}
\end{table}

% \begin{table}
% \centering
% \renewcommand*{\arraystretch}{1.5}
% \begin{tabular}{ p{0.9cm}|p{0.8cm}|p{0.75cm}|p{1cm}|p{1cm}|p{1cm}| } 
%  \cline{1-6} Time horizon ($s$) 
%  &  RMSE ($m$) &  Acc ($m/s^2$) & Yaw jerk ($rad/s^3$) & Pitch jerk ($rad/s^3$) & \review{Solution time} ($s$) \\ \hline\hline
%  0.5 &    1.21  & 0.15 & 10.68      & 1.35 & 0.004        \\ \hline
%  2 & 0.9 & 0.08 & 5.12  & 0.32 & 0.016 \\ \hline
%  4 		     &  1.23 & 0.05 &  5.06 & 0.32 & 0.029   \\ \hline
%  8 	     &  1.3 & 0.04 & 4.8 & 0.29 & 0.101 \\ \hline
% \end{tabular}
% \caption{Resulting metrics for the lateral shot. In this case, the baseline UAV trajectory is a straight line connecting the start and final positions. \emph{\review{Solution time}} is the average time to compute each trajectory.
% }\label{tab:metricsLateral}
% \end{table}

\begin{figure*}[h]
\centering
\includegraphics[width=1.65\columnwidth]{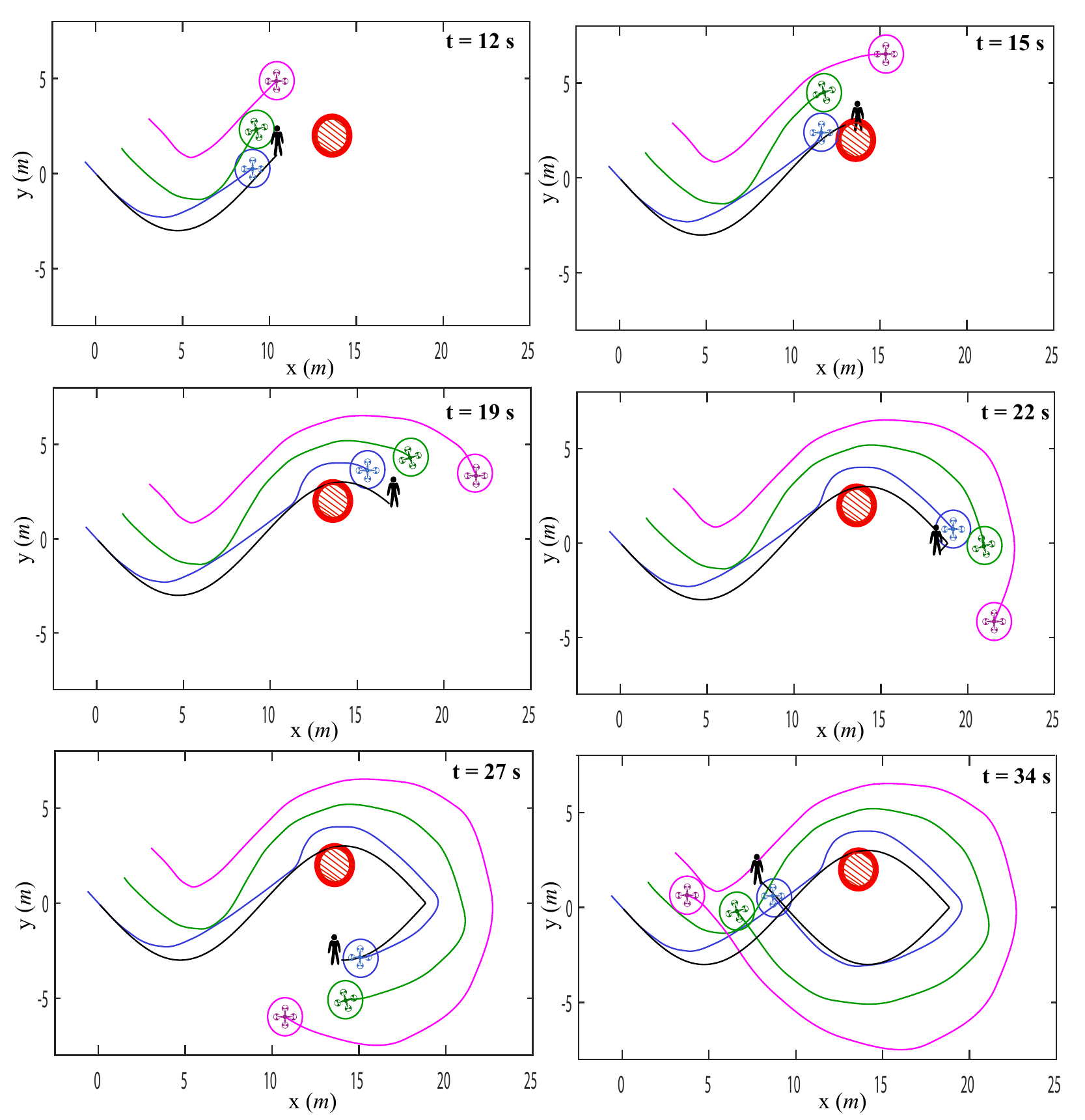}
\caption{Top view of different time instants of an experiment with three UAVs filming a target in a coordinated manner. The target (black) follows an 8-shaped path. UAV 1 (blue) performs a chase shot $2~m$ behind the target, UAV 2 (green) a lateral shot $2~m$ aside the target and UAV 3 (magenta) an orbit shot with a $4~m$ radius. A no-fly zone (red) is placed in the middle of the target trajectory for multi-UAV avoidance. Each UAV is represented with a $2~m$ circle around to show the collision avoidance constraint.}
\label{fig:multiUAV_experiment}
\end{figure*}

 \begin{figure}[htb]
    \centering
    \includegraphics[width=1\columnwidth]{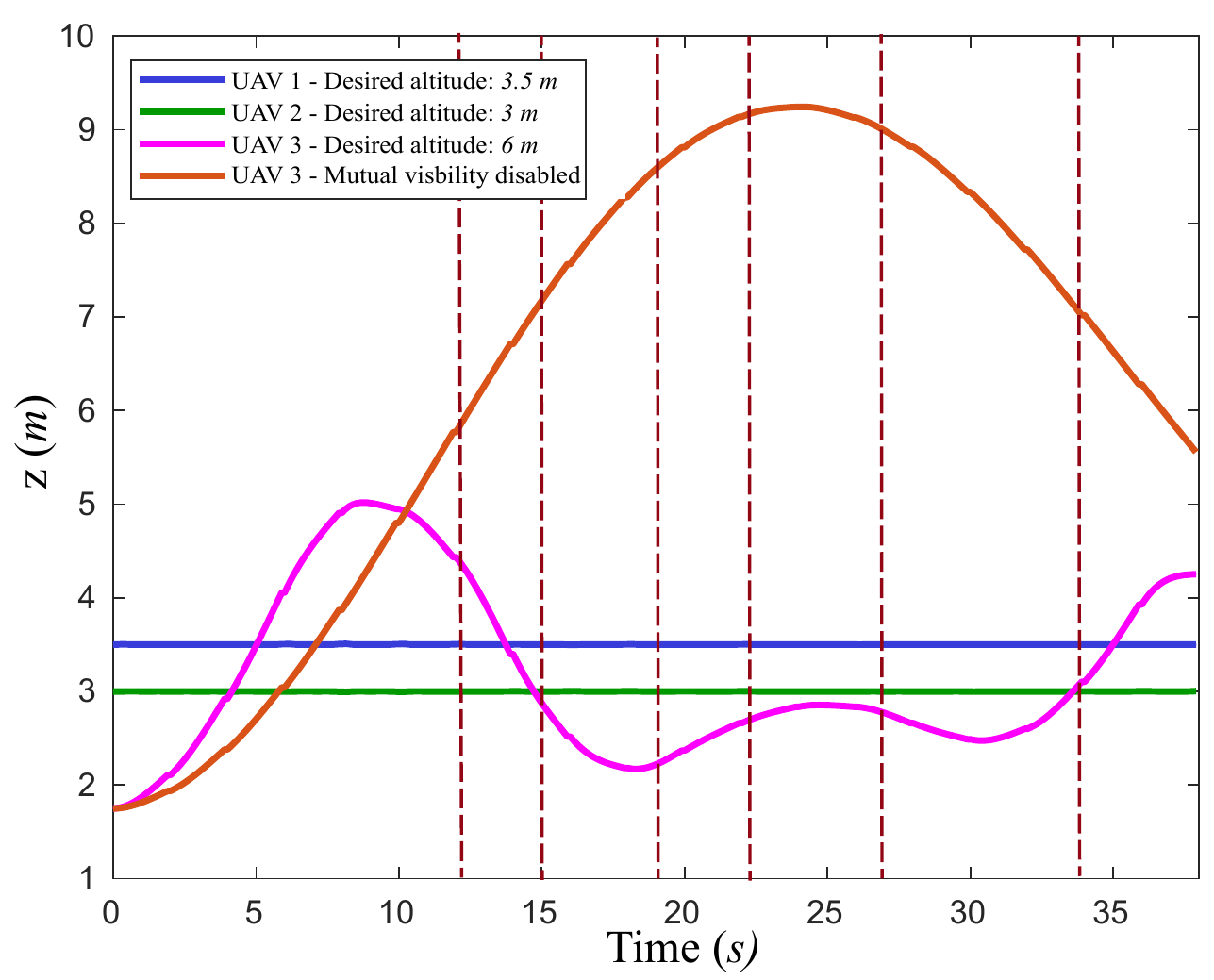}
    \caption{Temporal evolution of the UAV altitudes during the multi-UAV coordination experiment. The vertical lines mark the time instants of the snapshots depicted in Figure~\ref{fig:multiUAV_experiment}.}
    \label{fig:heights_multi_UAV}
\end{figure}

\subsection{Multi-UAV coordination}

In order to test multi-UAV coordination, we performed another experiment with three UAVs filming a target that followed an 8-shaped path at a speed of $1~m/s$. We combine heterogeneous shots from Type I and II, with a duration of 40 seconds each: UAV 1 performs a chase shot at a $3.5~m$ altitude and $2~m$ behind the target; UAV 2 a lateral shot at a $3~m$ altitude and $2~m$ as lateral distance from the target; and UAV 3 is commanded an orbit of radius $4~m$ at an altitude of $6~m$. Each UAV ran our method with a receding horizon of $N=40$ ($4~s$), recomputing trajectories at $0.5~Hz$. We set $r_{col}=2~m$ for collision avoidance and the low-pitch configuration for cost function weights, as we saw this was working better for this experiment. The purpose of this experiment is twofold. First, we show how the method works with a non-rectilinear target motion. We assume known the course of the \emph{road} followed by the target, and hence, we use a model that constrains the target motion to that path. Second, we show the main features related to multi-UAV coordination. A no-fly zone is used to test obstacle avoidance in a coordinated manner, including inter-UAV collision avoidance and mutual visibility avoidance.

Figure~\ref{fig:multiUAV_experiment} depicts several snapshots of the experiment~\footnote{For the sake of clarity, a video with the temporal evolution of the simulation is available at \url{https://youtu.be/u5Vi4leni7U}.}. We set UAV 1 as the one with top priority in the trajectory planner, then UAV 2, and least priority for UAV 3. Between $t=15~s$ and $t=19~s$, UAV 1 comes across the no-fly zone in its trajectory and deviates to avoid it. Consequently, UAV 2 also deviates in a coordinated manner not to collide with UAV 1. Although UAV 1 and 2 are close throughout the whole experiment, they keep a safety distance above $2~m$. UAV 3 is commanded an orbit from $4~m$, but between $t=19~s$ and $t=27~s$, it moves incrementally further from the target as it takes the orbit. This makes sense to minimize the variation in camera angles, as that UAV increases its altitude slightly during that period. Figure~\ref{fig:heights_multi_UAV} shows the temporal evolution of the UAV altitudes during the experiment. UAV 3 starts the experiment $1.5~m$ high and its commanded altitude for the orbit is $6~m$. We provided that desired altitude for the shot to enforce coordination, as we detected that at that altitude the other two UAVs where appearing within the field of view of UAV 3. UAV 1 and 2 start at their commanded altitudes and keep them throughout the entire experiment, as they have no issues with mutual visibility. However, UAV 3 starts ascending to reach its commanded altitude, which is never reached to comply with the mutual visibility constraint. As UAV 3 has less priority, it is the one changing its altitude during the experiment to avoid UAV 1 and 2 within its field of view. We also included in Figure~\ref{fig:heights_multi_UAV} the trajectory of UAV 3 altitude when the mutual visibility constraint is disabled in the planner. In that case, it can be seen that the UAV ascends above $6~m$, which causes the other two UAVs to appear within its field of view.

%%%%%%%%%%%%%%%%%%%%%%%%%%%%%%%%%%%%%%%%%%%
%%%%%%%%%%%%%%%%%%%%%%%%%%%%%%%%%%%%%%%%%%%
\section{Field Experiments}
\label{sec:fieldExp}

In this section, we report on field experiments to test our method with 3 UAVs filming a human actor outdoors. This allows us to verify the method feasibility with actual equipment for UAV cinematography and assess its performance in a scenario with uncertainties in target motion and detection.  

\begin{figure}[tbh]
    \centering
    \includegraphics[width=0.9\columnwidth]{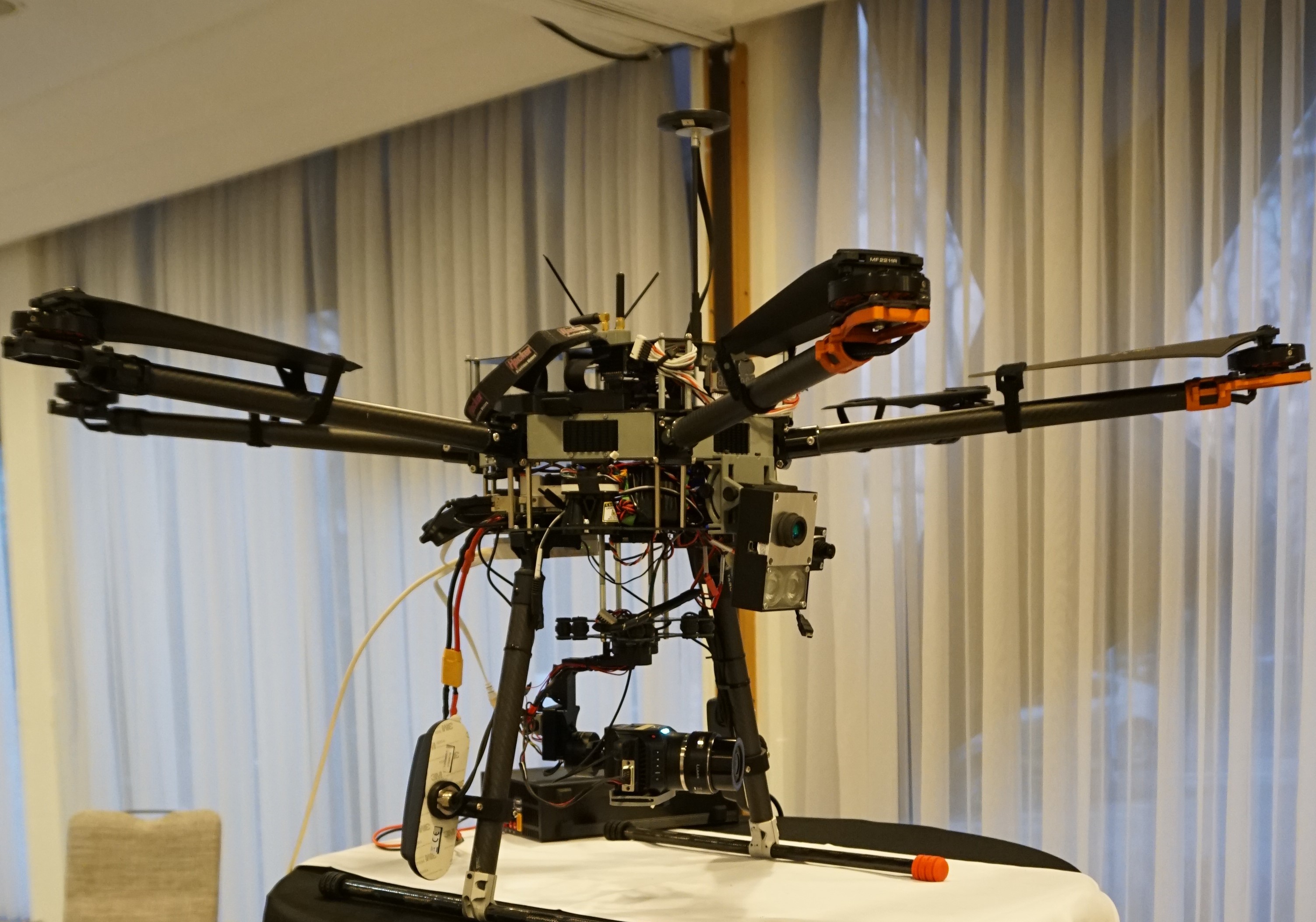}
    \caption{One of the UAVs used during the field experiments.}
    \label{fig:drone}
\end{figure}

The cinematography UAVs used in our experiments were like the one in Figure~\ref{fig:drone}. They were custom-designed hexacopters made of carbon fiber with a size of $1.80\times1.80\times0.70~m$ and had the following equipment: a PixHawk 2 autopilot running PX4 for flight control; an RTK-GPS for precise localization; a 3-axis gimbal controlled by a BaseCam (AlexMos) controller receiving angle rate commands; a Blackmagic Micro Cinema camera; and an Intel NUC i7 computer to run our software for shot execution. The UAVs used Wi-Fi technology to share among them their plans and communicate with our Ground Station. Moreover, as explained in Section~\ref{sec:system_architecture}, our target carried a GPS-based device during the experiments, to provide positioning measures to the Target Tracker component on board the UAVs. The device weighted around 400 grams and consisted of an RTK-GPS receiver with a Pixhawk, a radio link and a small battery. This target provided 3D measurements with a delay below $100~ms$, that were filtered by the Kalman Filter on the Target Tracker to achieve centimeter accuracy. These errors were compensated by our gimbal controller to track the target on the image.

{}

\begin{figure}[tbh]
    \centering
    \includegraphics[width=0.85\columnwidth]{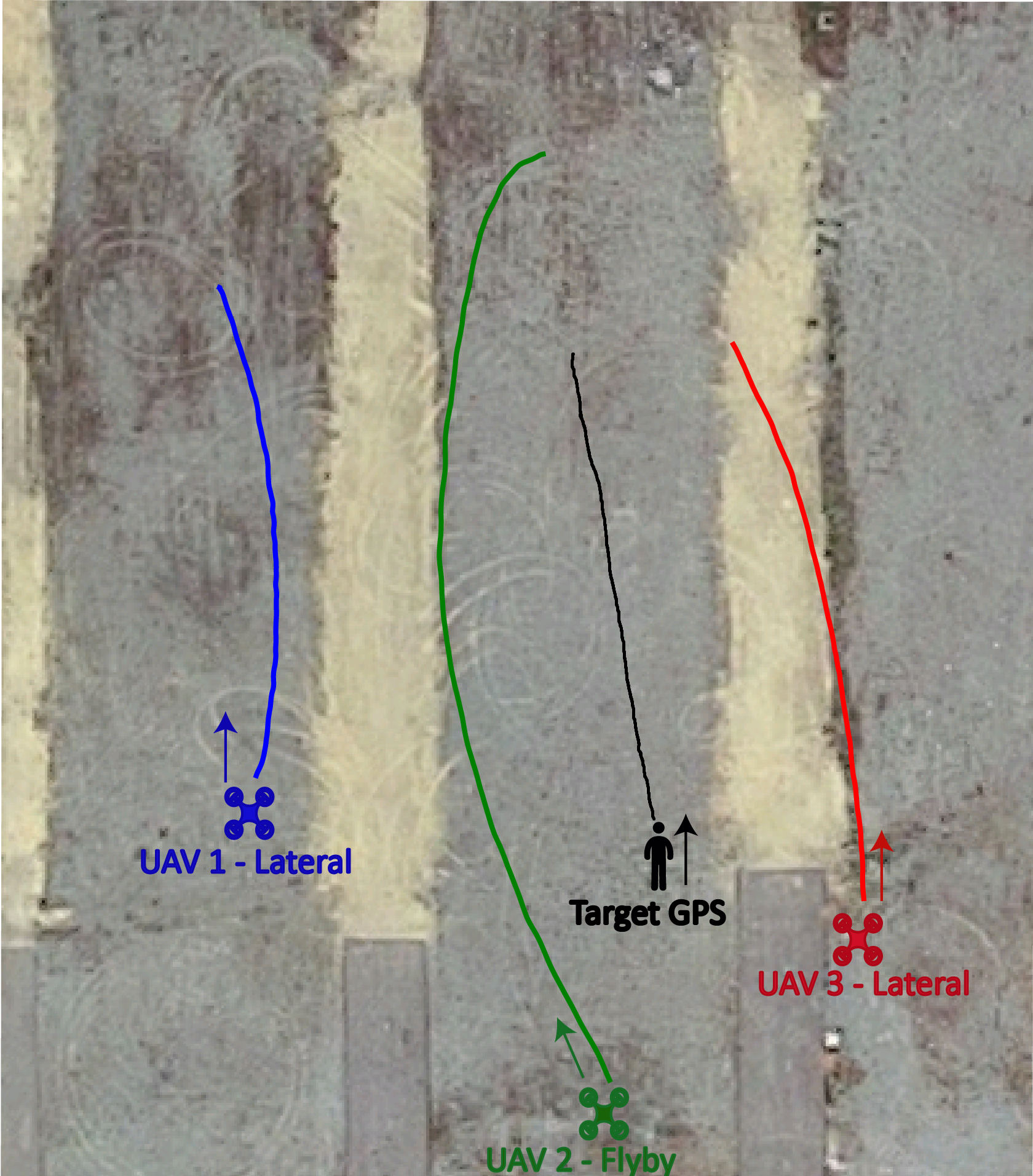}
    \caption{Trajectories followed by the UAVs and the human target during the field experiment. UAV 1 (blue) does a lateral, UAV 2 (green) a flyby and UAV 3 (red) a lateral.}
    \label{fig:trajectories_over_map}
\end{figure}{}

We integrated the architecture described in Section~\ref{sec:system_architecture} into our UAVs, using the ROS framework. We developed our method for trajectory planning (Section~\ref{sec:method}) in C++~\footnote{\url{https://github.com/alfalcmar/optimal_navigation}.}, using Forces Pro~\cite{FORCESnlp} to generate a compiled library for the non-linear optimization of our specific problem. The parameters used in the experiments were also those in Table~\ref{tab:parameters}. For collision avoidance, we used $r_{col}=5~m$, a value slightly increased for safety w.r.t. our simulations. We also limited the maximum velocity of the UAVs to $1~m/s$ for safety reasons.
% In these experiments, we used the high-pitch configuration for cost functions weights, as removing the term related with yaw was reducing non-linearities and producing a more robust behavior of the solver, which was capital in a real experiment.
All trajectories were computed on board the UAVs online at $0.5~Hz$, with a receding horizon of $N=100$ ($10~s$). Then, the Trajectory Follower modules generated 3D velocity commands at $10~Hz$ to be sent to the UAL component, which is an open-source software layer~\footnote{\url{https://github.com/grvcTeam/grvc-ual}.} developed by our lab to communicate with autopilot controllers. Moreover, we assumed a constant speed model for the target motion. This model was inaccurate, as the actual target speed was unknown, but those uncertainties were addressed by recomputing trajectories with the receding horizon. 

We designed a field experiment with 3 UAVs taking simultaneously different shots of a human target walking on the ground. UAV 1 performs a lateral shot following the target sideways with a lateral distance of $20~m$; UAV 2 performs a flyby shot starting $15~m$ behind the target and finishing $15~m$ ahead in the target motion line; and UAV 3 performs another lateral shot, but from the other side and with a lateral distance of $15~m$. For safety reasons, we established different altitudes for the UAVs, $3~m$, $10~m$ and $7~m$, respectively. 
%The parameters of the shots resulting from the configuration provided by the director can be found in Table~\ref{tab:parameters_SA}. 
In our decentralized trajectory planning scheme, UAV 1 had the top priority, followed by UAV 2 and then UAV 3. Moreover, in order to design the shots of the mission safely and with good aesthetic outputs, we created a realistic simulation in Gazebo with all our components integrated and a \emph{Software-In-The-Loop} approach for the UAVs (i.e., the actual PX4 software of the autopilots was run in the simulator).

%\begin{table}[] 
%\begin{tabular}{lllllllll}
%\hline
%\multicolumn{1}{|l|}{} & \multicolumn{1}{l|}{Type} & \multicolumn{1}{l|}{$x_{i}$} & \multicolumn{1}{l|}{$y_{i}$} &  \multicolumn{1}{l|}{$z_{i}$}  &  \multicolumn{1}{l|}{$x_{e}$} & \multicolumn{1}{l|}{$y_{e}$} &  \multicolumn{1}{l|}{$z_{e}$} &  \multicolumn{1}{l|}{Order}\\ \hline
%\multicolumn{1}{|l|}{UAV 1} & \multicolumn{1}{l|}{Lateral} & \multicolumn{1}{l|}{0} & \multicolumn{1}{l|}{20} & \multicolumn{1}{l|}{3}  
%& \multicolumn{1}{l|}{0} & \multicolumn{1}{l|}{20} &  \multicolumn{1}{l|}{3} &  \multicolumn{1}{l|}{1}\\ \hline
%\multicolumn{1}{|l|}{UAV 2} & \multicolumn{1}{l|}{Flyby} & \multicolumn{1}{l|}{-15} & \multicolumn{1}{l|}{0} & \multicolumn{1}{l|}{10} 
%& \multicolumn{1}{l|}{15} & \multicolumn{1}{l|}{0} &  \multicolumn{1}{l|}{10} &  \multicolumn{1}{l|}{2}\\ \hline
%\multicolumn{1}{|l|}{UAV 3} & \multicolumn{1}{l|}{Lateral} & \multicolumn{1}{l|}{0} & \multicolumn{1}{l|}{-15} & \multicolumn{1}{l|}{7}  
%& \multicolumn{1}{l|}{0} & \multicolumn{1}{l|}{-15} &  \multicolumn{1}{l|}{7} &  \multicolumn{1}{l|}{3}\\ \hline
%\end{tabular}
%\caption{Parameters of the shots $[x_{i} y_{i} z_{i}]$ is the initial relative position of the UAVs respect to the target. The final is $[x_{e} y_{e} z_{e}]$}\label{tab:parameters_SA}
%\end{table}

\begin{figure}[ht]
    \centering
    \includegraphics[width=\columnwidth]{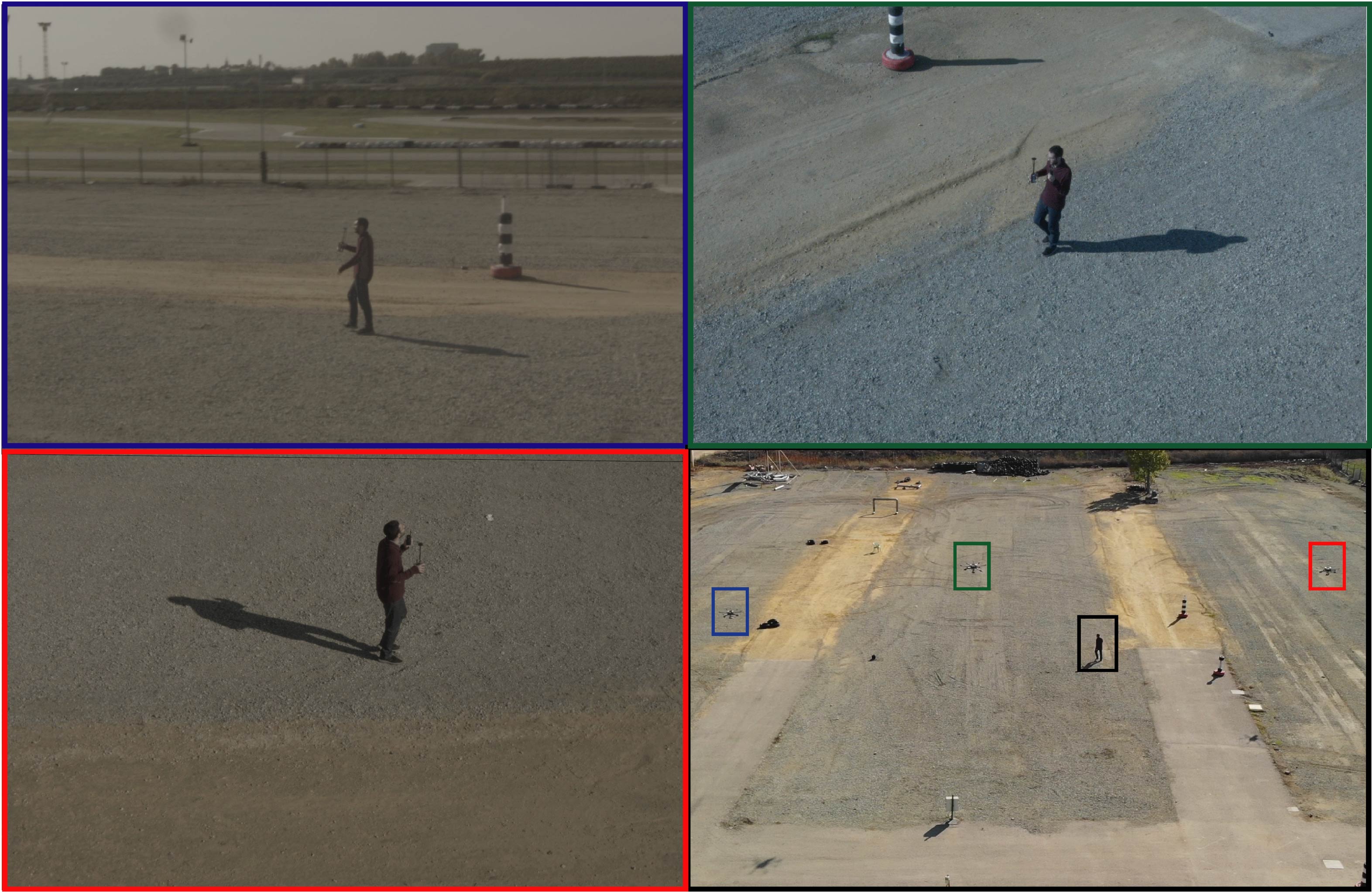}
    %\put(-332,210){\textcolor{blue}{1}}
    %\put(-163,210){\textcolor{green}{2}}
    \put(-332,100){\textcolor{red}{3}}
    % \put(-163,100){4}
    \put(-240,150){\textcolor{blue}{1}}
    \put(-120,150){\textcolor{green}{2}}
    \put(-240,68){\textcolor{red}{3}}
    \put(-80,40){Target}
    \caption{Example images from the cameras on board the UAVs during the experiment: top left, UAV 1 (blue); top right, UAV 2 (green); and bottom left, UAV 3 (red). Bottom right, a general view of the experiment with the three UAVs and the target.}
    \label{fig:cameras}
\end{figure}
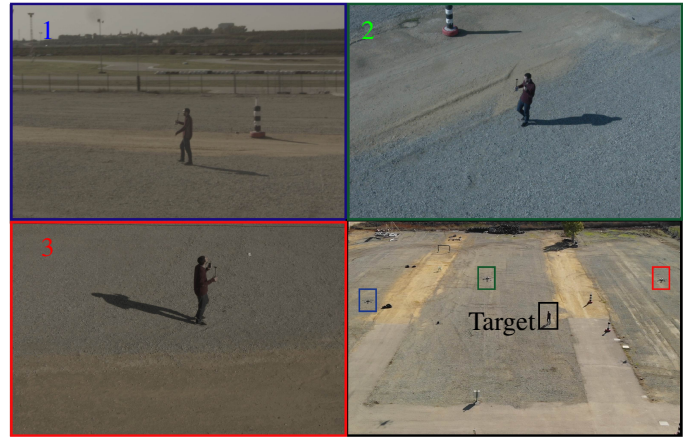{}

The full video of the field experiment can be found at \url{https://youtu.be/M71gYva-Z6M}, and the actual trajectories followed by the UAVs are depicted in Figure~\ref{fig:trajectories_over_map}. 
% The total duration of the mission was $47~s$. 
Figure~\ref{fig:cameras} shows some example images captured by the onboard cameras during the experiment. The experiment demonstrates that our method is able to generate online trajectories for the UAVs coping with cinematographic (i.e., no jerky motion, gimbal mechanical limitations and mutual visibility) and safety (i.e., inter-UAV collision avoidance) constraints; and keeping the target on the cameras' field of view, even under noisy target detections and uncertainties in its motion.  
Furthermore, we measured some metrics of the resulting trajectories (see Table~\ref{tb:metrics}) in order to evaluate the performance of our method. It can be seen that UAV accelerations were smooth in line with those produced in our simulations and the minimum distances between UAVs were always higher than the one imposed by the collision avoidance constraint ($5~m$).

\begin{table}
\centering
\renewcommand*{\arraystretch}{1.5}
\resizebox{0.7\columnwidth}{!}{%
\begin{tabular}{ |c|c|c|c| } 
 \cline{1-4} UAV & Traveled distance $(m)$ & Acc $(m/s^2)$ & Dist $(m)$\\ \hline\hline
1 & 95 & 0.125 & 9.543 \\ \hline 
2 & 141.4 & 0.108 & 9.543 \\ \hline 
3 & 98.6  & 0.100 & 14.711  \\ \hline
\end{tabular}}
\caption{Metrics of the trajectories followed by the UAVs during the field experiment. We measure the total traveled distance for each UAV, the average norm of the 3D accelerations and the minimum distance (horizontally) to other UAVs.}
\label{tb:metrics}
\end{table}

% \begin{table}[]
% \centering
% \begin{tabular}{|c|c|c|l|l|}
% \hline
% \multicolumn{1}{|r|}{\begin{tabular}[c]{@{}r@{}}UAV\end{tabular}} & \multicolumn{1}{l|}{\begin{tabular}[c]{@{}l@{}}Traveled\\ distance $(m)$ \end{tabular}} & \multicolumn{1}{r|}{Acc $(m/s^2)$} & Dist $(m)$ \\ \hline
% 1 & 95 & 0.125 & 9.543 \\ \hline 
% 2 & 141.4 & 0.108 & 9.543 \\ \hline 
% 3 & 98.6  & 0.100 & 14.711 \\ \hline
% \end{tabular}
% \caption{Metrics of the trajectories followed by the UAVs during the field experiment. We measure the total traveled distance for each UAV, the average norm of the 3D accelerations and the minimum distance (horizontally) to other UAVs.}
% \label{tb:metrics}
% \end{table}

\section{Conclusions}
\label{sec:conclusions}

In this paper, we presented a method for planning optimal trajectories with a team of UAVs in a cinematography application. We proposed a novel formulation for non-linear trajectory optimization, executed in a decentralized and online fashion. Our method integrates UAV dynamics and collision avoidance, as well as cinematographic aspects such as gimbal limits and mutual camera visibility. 
Our experimental results demonstrate that our method can produce coordinated multi-UAV trajectories that are smooth and reduce jerky movements. We also show that our method can be applied to different types of shots and compute trajectories online for time horizons of length up to 10 seconds, which seems enough for the considered cinematographic scenes outdoors. Moreover, our field experiments proved the applicability of the method with an actual team of UAV cinematographers filming outdoors.  

As future work, we plan to study alternative schemes for decentralized multi-UAV coordination instead of our priority-based computation. Our objective is to compute in a distributed manner multi-UAV approximate solutions that are closer to the optimum, but without increasing significantly the computation time, and test other team approaches like a leader-followers strategy. We believe that a comparison with methods based on reinforcement learning can also be of high interest.

\balance

\bibliographystyle{elsarticle-num-names}
\bibliography{optimal_trajectory_planning.bib}
\end{document}